\documentclass[sigconf]{acmart}

\usepackage{amsthm}
\usepackage{booktabs}
\usepackage{algorithm}
\usepackage{algpseudocode}
\usepackage{multirow}

\AtBeginDocument{%
  }

\setcopyright{acmlicensed}
\copyrightyear{2018}
\acmYear{2018}
\acmDOI{XXXXXXX.XXXXXXX}

\acmConference[Conference acronym 'XX]{Make sure to enter the correct
  conference title from your rights confirmation emai}{June 03--05,
  2018}{Woodstock, NY}
\acmISBN{978-1-4503-XXXX-X/18/06}

\begin{document}

\title{PIP: Prototypes-Injected Prompt for Federated Class Incremental Learning}


\author{Muhammad Anwar Ma'sum}
\orcid{0000-0002-9251-7781}
\authornotemark[1]
\affiliation{%
  \institution{University of South Australia}
  \city{Adelaide}
  % \state{SA}
  \country{Australia}
}
\email{masmy039@mymail.unisa.edu.au}

% \email{muhammad_anwar.masum@mymail.unisa.edu.au}
% \email{some.one@long. \linebreak university.edu}

\author{Mahardhika Pratama}
\orcid{0000-0001-6531-5087}
\affiliation{
  \institution{University of South Australia}
  \city{Adelaide}
  \country{Australia}}
\email{dhika.pratama@unisa.edu.au}

% \affiliation{%

\author{Savitha Ramasamy}
\orcid{0000-0003-1534-2989}
\affiliation{
  \institution{Institute of Infocomm Research}
  \city{Singapore}
  \country{Singapore}}
\email{ramasamysa@i2r.a-star.edu.sg}

\author{Lin Liu}
\orcid{0000-0003-2843-5738}
\affiliation{
  \institution{University of South Australia}
  \city{Adelaide}
  \country{Australia}}
\email{lin.liu@unisa.edu.au}

\author{Habibullah Habibullah}
\orcid{0000-0002-9542-9525}
\affiliation{
  \institution{University of South Australia}
  \city{Adelaide}
  \country{Australia}}
\email{habibullah.habibullah@unisa.edu.au}

\author{Ryszard Kowalczyk}
\orcid{0000-0003-0937-4028}
\affiliation{
  \institution{University of South Australia}
  \city{Adelaide}
  \country{Australia}}
\email{ryszard.kowalczyk@unisa.edu.au}

% \affiliation{%

\renewcommand{\shortauthors}{Author et al.}

\begin{abstract}
Federated Class Incremental Learning (FCIL) is a new direction in continual learning (CL) for addressing catastrophic forgetting and non-IID data distribution simultaneously. Existing FCIL methods call for high communication costs and exemplars from previous classes. We propose a novel rehearsal-free method for FCIL named prototypes-injected prompt (PIP) that involves 3 main ideas: a) prototype injection on prompt learning,  b) prototype augmentation, and c) weighted Gaussian aggregation on the server side.  Our experiment result shows that the proposed method outperforms the current state of the arts (SOTAs) with a significant improvement (up to $33\%$) in CIFAR100, MiniImageNet and TinyImageNet datasets. Our extensive analysis demonstrates the robustness of PIP in different task sizes, and the advantage of requiring smaller participating local clients, and smaller global rounds. For further study, source codes of PIP, baseline, and experimental logs are shared publicly in \textbf{\url{https://github.com/anwarmaxsum/PIP}}.
\end{abstract}




\received{20 February 2007}
\received[revised]{12 March 2009}
\received[accepted]{5 June 2009}

\maketitle

\section{Introduction}
% \vspace{-5pt}
Federated learning (FL) offers a collaborative approach for many clients to produce a shared global model while protecting their data privacy \cite{mcmahan2017communication}\cite{karimireddy2020scaffold} \cite{shoham2019overcoming}. FL has recently sparked a great deal of academic interest and achieved outstanding success in various application areas, including medical diagnosis \cite{hwang2023towards},  autonomous vehicle \cite{he2023three}, and wearable technology \cite{baucas2023federated}. However, the majority of FL methods are designed for a static application scenario, assuming the target classes are fixed and known in advance. In real-world applications, the data are dynamic, allowing local clients to access unseen target classes online.  

Existing studies have addressed dynamic data challenges in FL through Federated Class Incremental Learning (FCIL) where each local client gathers training data continually according to their continuous observation of the environment. In contrast, new clients with unforeseen classes are always welcome to join the FL training \cite{dong2023no} \cite{dong2022federated} \cite{yoon2021federated}.  The clients have to cooperatively train a global model that continually learns new classes while maintaining its capability to recognize the previous classes. Despite its capability to learn new classes, the global model tends to lose its knowledge of previous classes known as catastrophic forgetting. Therefore, FCIL calls for answers on how to handle catastrophic forgetting in a dynamic collaborative continual learning, while preserving data privacy. In addition, the clients may carry non-independently and identically distributed (non-i.i.d.) data with imbalance classes.

The current FCIL state-of-the-art (SOTA) i.e. GLFC\cite{dong2022federated} and LGA\cite{dong2023no}  train and share whole backbone parameters resulting in a large number of parameters (11.3M) to train that requires a longer training time, and a high communication cost (61.7MB)  as shown in Table \ref{tab:ours_vs_sota}. In addition, in the case of client-server limited communication costs e.g. due to limited bandwidth, this approach is forced to utilize a smaller (more shallow) model to reduce the shared parameter size. Second, in the current SOTAs, the clients privately share perturbed images with the central server for the aggregation process. Besides its inefficient size (16.5MB), sharing perturbed images may violate data privacy principles since perturbation doesn't guarantee information leakage. Furthermore, this mechanism can't answer the data openness problem, where the data is only open for a client at a specific moment. These intriguing gaps motivate us to develop a more efficient yet more effective approach to the FCIL problem, where a client trains and shares as small parameters as possible but produces a highly accurate global model without sharing any (perturbed) samples.
\begin{table}[h]
% \vspace{-8pt}
\centering
% \scriptsize
\small
\setlength{\tabcolsep}{0.2em}
\renewcommand{\arraystretch}{0.5}
\begin{tabular}{l|c|ccc|cc|c}
\hline
\multirow{2}{*}{Method}                                   & \multirow{2}{*}{B.Bone} & \multicolumn{3}{c}{Trainable Params} & \multicolumn{2}{c}{Prototype}                                                           & \multirow{2}{*}{\begin{tabular}[c]{@{}c@{}}Com.\\ Cost(B)\end{tabular}} \\ \cline{3-7}
                                                          &                         & Type       & \#N       & Size(B)     & Type                                                                          & Size(B) &                                                                         \\ \hline
GLFC                                                      & ResNet18*                     & ResNet18*        & 11.3M     & 45.2 M      & \begin{tabular}[c]{@{}c@{}}Perturbed\\ Images*\end{tabular} & 16.5 M  & 61.7 M                                                                  \\
LGA                                                       & ResNet18*                     & ResNet18*        & 11.3M     & 45.2 M      &  \begin{tabular}[c]{@{}c@{}}Perturbed\\ Images*\end{tabular}                                                                              & 16.5 M  & 61.7 M                                                                  \\ \hline
\begin{tabular}[c]{@{}l@{}}Baseline\\ (Ours)\end{tabular} & ViT$^x$                     & Prompt*     & 0.33M     & 1.32 M      & -                                                                             & -       & 1.32 M                                                                  \\
\begin{tabular}[c]{@{}l@{}}PIP\\ (Ours)\end{tabular}      & ViT$^x$                      & Prompt*     & 0.33M     & 1.32 M      & \begin{tabular}[c]{@{}c@{}}Deep \\ Features*\end{tabular}                      & 0.03 M  & 1.35 M  \\ \hline                                                            
\end{tabular}
% \fontsize{4}{4}
\caption{\footnotesize The difference between our proposed approach and current SOTA in Federated Class Incremental Learning (FCIL), * indicates sharable parameters, while $^x$ indicates unsharable parameters.}
\label{tab:ours_vs_sota}
\end{table}

In this study, we propose a new approach named federated prompt for the FCIL problem that is implemented in our proposed baselines and our proposed method named prototype-injected prompt (PIP).  In our proposed approach, a client trains and shares only a small (0.33M) set of parameters called prompt, while the backbone model remains frozen and unsharable. In our approach, the bound for communication cost is only in the prompt size not in the backbone size. As an advantage, we could utilize any backbone with any complexity, number of parameters, and size. Our approach utilizes deep features as a shareable prototype to handle the non-i.i.d data between clients. As the second advantage, our approach offers far smaller prototypes (0.03MB) and guarantees data privacy as deep features represent a totally different structure and value compared to raw or perturbed samples. The key points difference between our approach and current SOTA is described in Table \ref{tab:ours_vs_sota}.
\textbf{The contributions} of this paper are summarized as follows:
 \begin{itemize}
 % \vspace{-5pt}
\item We propose a new approach, prompt-based federated continual learning to solve the FCIL problem, where clients and central servers exchange prompts and head layer units instead of a whole model. This approach significantly reduces the size of the exchanged parameters between clients and the central server in each round.
% \vspace{-5pt}
\item We propose a new baseline method for the FCIL problem named Federated DualPrompt (Fed-DualP) that already outperforms the current state-of-the-art (SOTA) methods in three benchmark datasets.
% \vspace{-5pt}
\item We propose a novel method for the FCIL problem named Prototype-Injected Prompt (PIP) that consists of three main ideas i.e. a) prototype injection on prompt learning,  b) prototype augmentation, and c) weighted Gaussian aggregation on the server side. The proposed method outperforms the baseline and current SOTAs with a significant gap.
% \vspace{-5pt}
\item We provide a comprehensive analysis in three benchmark datasets as well as the robustness of the proposed method in different task sizes, smaller participating clients, and smaller rounds per task. 
 % \vspace{-5pt}
\item The source codes of PIP are shared in \url{https://anonymous.4open.science/r/an122pouyyt789/} during the peer-review process and will be made public upon acceptance of our paper for further study and reproducibility.
% \vspace{-5pt}
 \end{itemize}
% \vspace{-9pt}

% \textbf{The contributions} of this paper are summarized as follows:
% (1) We propose a new approach, prompt-based federated continual learning to solve the FCIL problem, where clients and central servers exchange prompts and head layer units instead of a whole model. This approach significantly reduces the size of the exchanged parameters between clients and the central server in each round.
% (2) We propose a new baseline method for the FCIL problem named Federated DualPrompt (Fed-DualP) that already outperforms the current state-of-the-art (SOTA) methods in three benchmark datasets.
% (3) We propose a novel method for the FCIL problem named Prototype-Injected Prompt (PIP) that consists of three main ideas i.e. a) prototype injection on prompt learning,  b) prototype augmentation, and c) weighted Gaussian aggregation on the server side. The proposed method outperforms the baseline and current SOTAs with a significant gap.
% (4) We provide a comprehensive analysis in three benchmark datasets as well as the robustness of the proposed method in different task sizes, smaller participating clients, and smaller rounds per task. 
% (5) The source codes of PIP are shared in \url{https://anonymous.4open.science/r/an122pouyyt789/} during the peer-review process and will be made public upon acceptance of our paper for further study and reproducibility.

\section{Related Works}
% \vspace{-5pt}
\textbf{a). Class Incremental Learning (CIL):} Prompt-based approach e.g. L2P \cite{wang2022learning}, DualPrompt \cite{wang2022dualprompt}, CODA-Prompt \cite{smith2023coda} offers a breakthrough solution for CIL by training only small amounts of parameters called \textbf{prompt} for down-streaming tasks sequence, while the backbone model which contains the biggest parameter numbers stays frozen. This approach reduces the training time and offers utilization of a more complex backbone model e.g. ViT instead of CNN. The rehearsal approach e.g. ICARL \cite{rebuffi2017icarl}, EEIL, \cite{castro2018end}, GD \cite{prabhu2020gdumb}, DER++ \cite{buzzega2020dark},  trains exemplars (memory) from the previous task joined with current task samples to minimize model forgetting on previous tasks. This approach doesn't work well if the memory samples are not available. The bias correction approach e.g. BiC \cite{wu2019large} and LUCIR \cite{hou2019learning} enhances the rehearsal approach by creating a task-wise bias layer to help the model achieve stability-plasticity balance in compensation to add extra parameters to be trained. The regularization approaches adaptively tune the base learner parameter to accommodate the previous task and current task e.g. EWC  \cite{kirkpatrick2017overcoming}, MAS \cite{aljundi2018memory}, LWF \cite{li2017learning}, DMC \cite{zhang2020class}. The regularization has the advantage that It works on memory-free CIL but on the contrary, it is outperformed by rehearsal and bias correction approaches if the memory is available. Prompt-based approach outperforms the other three approaches despite requiring no memory, but not yet been proven effective in a distributed (federated) setting.

\noindent\textbf{b). Federated Class Incremental Learning(FCIL):} The recent study on FCIL problem e.g. FedWeIT\cite{yoon2021federated}, GLFC \cite{dong2022federated}, and LGA \cite{dong2023no} strives to achieve an optimal global model by aggregating locally trained models sent by the clients. The current SOTAs are proven to be more effective than combining FedAvg \cite{mcmahan2017communication} and class incremental learning approaches such as ICARL and BiC. However, as emphasized in Table \ref{tab:ours_vs_sota}, the current SOTAs train and share all backbone parameters, resulting in a less efficient training time and high communication costs. In Addition, the SOTAs assume that a client saves several exemplars as local memory and shares perturbated images with the server which is not always applicable in real applications. The other study i.e. TARGET \cite{zhang2023addressing} utilizes a synthetic dataset instead of saving exemplars from the previous tasks to train the whole backbone network. TARGET is proven to be more effective than Fed-LWF, Fed-EWC, and FedWeIT\cite{yoon2021federated} but still outperformed by LGA and GLFC. Another approach i.e. FedCIL \cite{qi2023better} uses a locally trained generative model i.e. ACGAN \cite{odena2017conditional} to generate fake samples in the aggregation process. Despite its better performance than the combination of FedAvg or FedProx \cite{li2020federated} with ACGAN, DGR\cite{shin2017continual} or LWF-2T \cite{usmanova2021distillation} this method requires higher training time both in local and server-side and higher communication cost due to sending the generative model along with the backbone. To tackle the weaknesses above, our approach uses memory-free prompt learning for local training with the frozen backbone to achieve an efficient yet powerful local model, then aggregates the prompt and prototypes to obtain a generalized global model with low communication cost.
% \vspace{-9pt}

\begin{figure*}[h!]
\begin{center}
%\framebox[4.0in]{$\;$}
\includegraphics[width=0.95\textwidth]{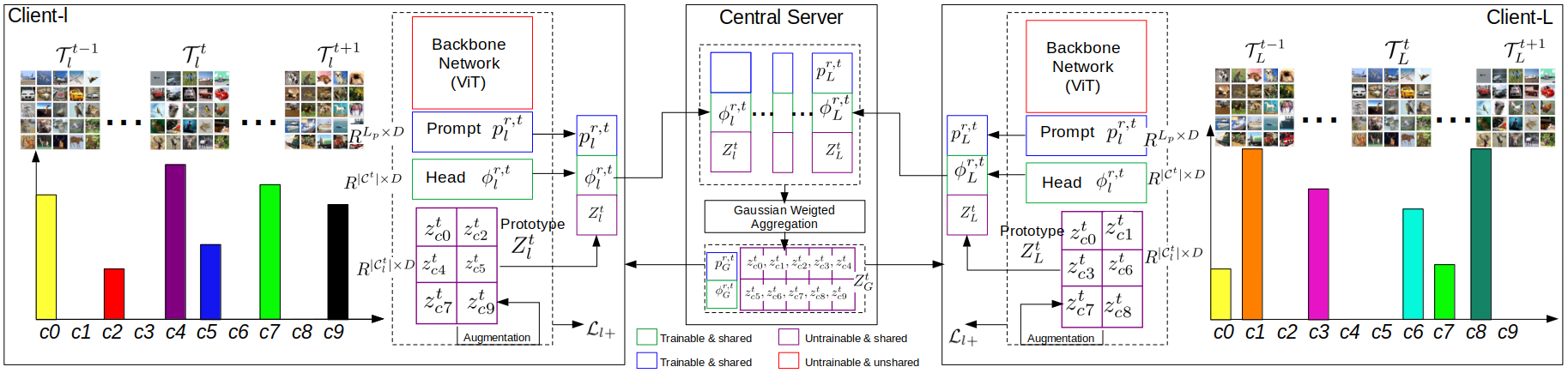}
\end{center}

\caption{Visualization of our proposed method (PIP) that includes prototypes-injected-prompt learning with $\mathcal{L}_{l+}$ to handle local catastrophic forgetting, shared prototypes to handle non-i.i.d distributions between clients, prototypes augmentation to handle class-imbalance, and server Gaussian weighted aggregation to improve global model generalizations.  }
\label{fig:pip_diagram}
\end{figure*}
% \vspace{-2pt}
\section{Preliminaries} 
% \vspace{-2pt}
\subsection{Problem Formulation}
% \vspace{-2pt}
\textbf{Class incremental learning (CIL)} problem is defined as a learning problem of a sequence of fully supervised learning tasks $\{\mathcal{T}^t\}_{t=1}^T$ where $T$ represents the number of consecutive tasks, the value of $T$ is unknown to the model before CIL process and can be infinite. Each task carries $N_t$ pairs of training samples $\mathcal{T}^{t}=\{x_i^{t},y_i^{t}\}_{i=1}^{N_t}$ where $x_i\in\mathcal{X}^t$ denotes input image and $y_i\in\mathcal{Y}^t$ denotes its corresponding class label. Each task carries the same image size but possesses disjoint target classes to the other tasks. Suppose that $\mathcal{Y}^t$ is the labels set of task $t$ and $\mathcal{C}^{t}$ denotes the unique class labels in $\mathcal{Y}^t$, $\mathcal{C}^{t}=\{y^t_i,y^t_j \in \mathcal{Y}^t, i,j \in [1..N^t]\}_{\neq}$, then we have  $\forall t, t' \in[1..T], t\neq t' \Rightarrow (\mathcal{Y}^t\cap \mathcal{Y}^{t'}=\emptyset) \wedge (\mathcal{C}^t\cap \mathcal{C}^{t'}=\emptyset$).

\noindent\textbf{Federated Class-Incremental Learning (FCIL)} is defined as:  for $r-th$ global round $r \in [1..R]$, a set of selected local clients $\{S_l\}^L_{l=1}$ are conducting class incremental learning for the task $t \in [1..T]$ with their local training set $\{\mathcal{T}^{t}_{l}\}_{l=1}^{L}$ coordinated by a central server $S_G$ in a federated manner, where $R$ is the maximum global round. A client $S_l$ may carry a unique set of a training set $\mathcal{T}_{l}^{t}=\{x_{li}^{t},y_{li}^{t}\}_{i=1}^{N^{t}_l}$, which satisfies $\forall l, l' \in[1..L], l\neq l' \Rightarrow (\mathcal{T}^{t}_{l} \neq \mathcal{T}^{t}_{l'}) \wedge (\mathcal{C}^{t}_{l} \neq \mathcal{C}^{t}_{l'})$ where, ${N^{t}_l}$ denotes number of training samples of client $l$ for task $t$. FCIL satisfies $\mathcal{T}^{t} = \cup_{l=1}^{L} \mathcal{T}^{t}_{l}$ and $\mathcal{C}^{t} = \cup_{l=1}^{L} \mathcal{C}^{t}_{l}$, therefore it satisfies $\mathcal{C}^{t}_l \subseteq \mathcal{C}^{t}$ and $|\mathcal{C}^{t}_l| \leq |\mathcal{C}^{t}|$.  Please note that $\mathcal{T}^{t}_{l}$ is not shareable to another client $S_{l'} (l\neq l')$  nor central server $S_G$. Clients may carry non-identically and distributed data (non-i.i.d), suppose that $\mathcal{D}_l$ is the class distribution of $l-th$ client, then it satisfies $\forall l, l' \in[1..L], l\neq l' \Rightarrow (\mathcal{D}^{t}_{l} \neq \mathcal{D}^{t}_{l'})$. In each global round $r-th$ for task $t$,  each client $S_l$ conducts local CIL training using its training samples $\mathcal{T}^{t}_l$ optimizing its local model $\Theta^{r,t}_l$, then the clients send their local models to the central server $S_G$. The central server aggregates all participating local updated parameters $\{\Theta^{r,t}_l\}_{l=1}^L$, producing the latest global model parameter $\Theta^{r,t}_G$, then distributes it to all clients $\{S_l\}^L_{l=1}$ for the next global round process. $\Theta^{r,t}_l$ and  $\Theta^{r,t}_g$ denote the latest updated version of local parameter $\Theta_l$ and global parameter $\Theta_g$ respectively after $r-th$ round of $t-th$ task of FCIL training. Please note that a set of local clients $\{S_l\}^L_{l=1}$ are randomly selected from all possibly participating clients simulating real-world conditions where only a small percentage of registered clients join in a round of federated learning.
% \vspace{-9pt}

\subsection{Proposed Approach and Baselines: Federated Prompt-Based Continual Learning for FCIL}
% \vspace{-3pt}
In prompt-based class incremental learning, a client uses a frozen backbone model e.g. Vision Transformer (ViT) as a feature extractor. The ViT embedding layer transforms an input image $x \in R^{w \times h \times c}$, where $w,h,c$ are the image's width, height, and channel respectively, into sequence-like output feature $h \in R^{L \times D}$, where $L$ is the sequence length and $D$ is the embedding dimension. A client only needs to update a small-sized trainable parameter called prompt $p \in R^{L_p \times D}$ and its head layer parameter ${\phi}$ to deal with the sequence of task $\{\mathcal{T}_t\}_{t=1}^T$, $L_p$ denotes the prompt length. Inspired by its effectiveness and efficiency in CIL problem, We proposed a federated prompt-based continual learning for the FCIL problem, where in a global round $r-th$ for the $t-th$ task a client $S_l$ only updates its learnable parameters i.e. prompt $p_l^{r,t} \in R^{L_p \times D}$,  and its classification (head) layer $\phi_l^{r,t}$, then send them to central server $S_G$. The central $S_G$ aggregates accumulated parameters $\{(p_l^{r,t},\phi_l^{r,t})\}_{l=1}^L$ producing global model parameters $(p_G^{r,t}, \phi_G^{r,t})$ and send it back to all clients. The size of $(p_l^{r,t}, \phi_l^{r,t})$ is far smaller than the whole model $\Theta_l = (\theta_l, \phi_l)$.  Therefore our proposed approach is more efficient both in terms of local training process and communication costs. The main distinction between our proposed approach and the existing state-of-the-art (SOTA) methods is that the SOTA methods train the whole model parameters $\Theta = (\theta, \phi)$ where a model is defined as $F_\Theta(x) = f_\phi(g_\theta(x))$, $g_\theta(.)$ is feature extractor and $f_\phi(.)$ is classifier, while our approach trains only small-sized parameters i.e. prompt $p$ and head layer $\phi$, and freezes the backbone $\theta$.

In this study, we propose two baseline methods i.e.  Federated Learning to Prompt (Fed-L2P), and  Federated DualPrompt (Fed-DualP) by customizing L2P \cite{wang2022learning}, and DualP \cite{wang2022dualprompt}, for federated class incremental learning problems. The main role of the baselines is to demonstrate that our new approach is more effective and efficient than current SOTAs for FCIL problem. Second, the baselines show that our approach applies to any prompt structure that is demonstrated by two different prompt structures i.e. L2P and DualP. In addition, we proposed those two methods as new baselines for further study on FCIL problem. Following L2P and DualP structure, in our baselines, the local client's prompt $p_{l}$ represented as $p_l=\{(e_l^t,k_l^t)\}^T_{t=1}$ in Fed-L2P $p_l=(g_l,\{(e_l^t,k_l^t)\}^T_{t=1})$ in Fed-DualP, where $p_l^t\in R^{L_p \times D}$ and $e_l^t\in R^{L_p \times D}$ represent a task-specific prompt, $g_l\in R^{L_p \times D}$ represents global prompt, $k_l^t$ is the key associated with a task-specific prompt $p_l^t$ or $e_l^t$, and $T$ denotes the number of tasks of client $S_l$. The central server aggregates participating clients' parameters by performing FedAvg \cite{mcmahan2017communication}. 
% \vspace{-10pt}
% Our baseline is formulated as: 
% \textbf{Prompt-Structure:} Following DualPrompt structure Fed-DualPrompt uses 2 prompt types, i.e. G-Prompt and E-Prompt. G-Prompt is a trainable shared parameter for all tasks defined as $g \in R^{L_g \times D}$ where $L_g$ is the sequence length and $D$ is the embedding dimension. E-Prompt is a set of task-specific parameters defined as $E = \{e^t\}_{t=1}^{T}$ where $e^t \in R^{L_e \times D}$, $L_e$ is the sequence length and $D$ is the embedding dimension.

\section{Proposed Method: Prototype-Injected Prompt (PIP)}\label{proposed_method}
% \vspace{-5pt}
Our proposed method, as visualized in Figure \ref{fig:pip_diagram} performs prompt learning on each client to handle local catastrophic forgetting (section \ref{local_prompt_learning}), handling non-i.i.d distribution by applying shared prototype injection(section \ref{prototype_injection}), handling class-imbalance by performing prototypes augmentation (section \ref{prototype_augmentaion}), and finally improving global model generalization by using weighted aggregation (section \ref{weighted_aggregation}).

\textbf{The uniqueness} of our method is that we utilize prompt and prototype as the shared knowledge between clients and the central server. Second, we propose a prototype injection mechanism in local training to improve local model generalization. Third, we propose a weighted aggregation on the server side. Our proposed method is distinct from prompt-based methods e.g. Fed-CPrompt\cite{bagwe2023fed} and FCILPT\cite{liu2023federated} that shared and trained prompts only. Our proposed method is also distinguished from prototype-based FL methods e.g. FedProto \cite{tan2022fedproto} and CCVR \cite{luo2021no} shared only prototypes to refine the local model or use the prototype as classifiers.  Besides, FedProto and CCVR are developed for federated learning only, meaning the methods didn't address the catastrophic forgetting issue.

% \vspace{-7pt}

\subsection{Handling Local Catastrophic Forgetting via Prompt Learning}\label{local_prompt_learning}
% \vspace{-5pt}
On a round $r-th$ of $t-th$ task, each client $S_l$ optimizes its local prompt $p_{l}$ and head layer parameter $\phi_l$ using its available training samples $\mathcal{T}_{l}^{r,t}$. Please note that $p_l$ can be any prompt structure as mentioned in our baselines e.g.in L2P, DualPrompt, or another prompt structure. Given a pre-trained ViT backbone $f$ with $M$ Multi head Self-Attention (MSA), where $h^{(i)}, i=1,2...,M$ represents the input for $i-th$ MSA layer, suppose that a local client wants to attach a latest updated prompt $p_l^{r,t}$ into the $i-th$ MSA layer, the prompt instance $p_l^{r,t}$ transform feature $h^{(i)}$ via prompt function as defined in equation \ref{prompting}.
 \begin{equation}\label{prompting}
 \small
        h_{p^{r,t}_l}^{(i)} = f_{prompt}(p^{r,t}_l,h^{(i)})
 \end{equation}
Note that $h_i$ is the extension of $h$, a sequence-like parameter produced by the ViT embedding layer for the $i-th$ MSA layer. Prompt function $f_{prompt}$ can be implemented by using prompt tuning ($f_{prompt}^{Pro-T}$) \cite{lester2021power} or prefix ($f_{prompt}^{Pre-T}$) tuning \cite{li2021prefix} approaches as defined in equation \ref{fpro} and \ref{fpre}.
\begin{equation}\label{fpro}
\small
    f_{prompt}^{Pro-T}(p^{r,t}_l,h^{(i)})=MSA([p^{r,t}_{l}{\oplus}h_Q^{(i)}],[p^{r,t}_{l}{\oplus}h_K^{(i)}],[p^{r,t}_{l}{\oplus}h_V^{(i)}])
\end{equation}
% \vspace{-12pt}
\begin{equation}\label{fpre}
\small
    f_{prompt}^{Pre-T}(p^{r,t}_l,h^{(i)})=MSA(h_Q^{(i)},[p^{r,t}_{lK}{\oplus}h_K^{(i)}],[p^{r,t}_{lV}{\oplus}h_V^{(i)}])
\end{equation}
\begin{equation}\label{msa}
    \small
    \begin{split}
        &MSA(h_Q^{(i)},h_K^{(i)},h_V^{(i)})=(h_1^{(i)}{\oplus}h_2^{(i)}{\oplus}...h_{m-1}^{(i)}{\oplus}h_m^{(i)})W^O\\
        &where \;  h_j^{(i)} = Attention(h_Q^{(i)}W_j^Q,h_K^{(i)}W_j^K,h_V^{(i)}W_j^V) 
    \end{split}
 \end{equation}
MSA function is defined as in equation \ref{msa} following \cite{vaswani2017attention} where  $h_Q^{(i)}$, $h_K^{(i)}$, and $h_V^{(i)}$ are input query, key, and value, for $i-th$ MSA layer, $p^{r,t}_{lK} \in R^{L_{p/2} \times D}$ and $p^{r,t}_{lV} \in R^{L_{p/2} \times D}$ are splits of $p^{r,t}_{l}$,  $W^O$, $W_j^Q$, $W_j^K$, and $W_j^V$ are the projection matrices, and $m$ is number of head, $\oplus$ represents concatenation function, and in ViT $h_Q^{(i)}$ = $h_K^{(i)}$ = $h_V^{(i)}$ = $h^{(i)} \in R^{L\times D}$. Given a pair of samples $(x_{li}^{t},y_{li}^{t})\in \mathcal{T}_l^{t}$, each client $S_l$ trains its latest updated parameters $p^{r,t}_l$ and $\phi^{r,t}_l$ using  $\mathcal{L}_l$ as defined in equation \ref{loss_client},
where $\mathcal{L}$ represents cross-entropy loss, $\mathcal{L}_{m}$ represents matching loss between sample $x_{li}^{t}$ and key $x_{li}^{t}$, $\lambda$ represents a constant factor, $f_{p_l}$ represents prompt function, and $f_{\phi_l}$ represents of $l-th$ client. 
\begin{equation}\label{loss_client}
\small
     \mathcal{L}_{l} = \mathcal{L}(f_{\phi_{l}^{r,t}}(f_{p_l^{r,t}}(x_{li}^{t}),y_{li}^{t})) + \lambda \mathcal{L}_{m}(x_l^{t},k_l^{r,t}), (x_{li}^{t},y_{li}^{t})\in \mathcal{T}_l^{t}
\end{equation}
% \vspace{-14pt}
\begin{equation}\label{param_update}
\small
\begin{split}
    &p_l^{r,t} \gets p_l^{r,t} - \alpha \nabla \mathcal{L}_l\\
    &\phi_l^{r,t} \gets \phi_l^{r,t} - \alpha \nabla \mathcal{L}_l 
\end{split}
\end{equation}
% \vspace{-2pt}
The client's parameters are updated using equation \ref{param_update} where $\alpha$ represents learning rate and $\nabla \mathcal{L}_l$ represents gradient with respect to $\mathcal{L}_l$.
% \begin{equation}\label{obj_client}
%     {p_l^*,\phi_l^*} = \arg \min_{p_l,\phi_l}{\mathbb{E}_{(x,y)\sim \mathcal{T}_l^{r,t}}\mathcal{L}_l(f_{\phi_l}(f_{p_l}(x),y))}
% \end{equation}
 
% \vspace{-7pt}
\subsection{Shared Prototype Injection: What, Why and How?}\label{prototype_injection}
% \vspace{-5pt}
We define a prototype set on the $t-th$ task of a client $S_l$ as $Z_l^t = \{z_{lc}^t\}, z_{lc}^t \in R^{1\times D}$ is the prototype for class $c \in \mathcal{C}_l^t$,  $\mathcal{C}_l^t$ is the available classes in $\mathcal{T}^{t}_l$, and $D$ is the embedding dimension. Assuming that the prototype follows a Gaussian distribution $z_{lc}^t \sim \mathcal{N}(\mu_{lc}^t,\,\Sigma_{lc}^t)$ and the prototype is considered as $D$ disjoint uni-variate distribution, then we have  $z_{lc}^t \sim \mathcal{N}(\mu_{lc}^t,\,{\sigma_{lc}^t}^{2})$  where ${\sigma_{lc}^t}^{2} = I_D.{\sigma_{lc,i}^t}^{2}, i \in \{1,2..D\}$, $I_D$ is identity matrix. Suppose that $\mathcal{T}_{lc}^t = \{(x_{li}^t, y_{li}^t) \in \mathcal{T}_l^t, y_{li}^t = c \}$ is the samples of class-$c$ in $\mathcal{T}_l^t$ and $|\mathcal{T}_{lc}^t|$ is the number of samples in $\mathcal{T}_{lc}^t$, we compute the prototype  $z_{lc}^t$  properties by Eq. \ref{proto_mu} and \ref{proto_sigma}.
% \vspace{-10pt}
\begin{equation}\label{proto_mu}
\small
    % \mu_{lc}^t =  \frac{1}{n^t_{lc}} \sum_{li=1}^{n^t_{lc}}f_{p_l^t}(x^t_{lci})\:  , x_{lci} \in \mathcal{T}_{lc}^t
    \mu_{lc}^t =  \frac{1}{|\mathcal{T}^t_{lc}|} \sum_{li=1}^{|\mathcal{T}^t_{lc}|} f_{p_l^{r,t}}(x^t_{lci})\:  , x_{lci} \in \mathcal{T}_{lc}^t
\end{equation} 
\vspace{-7pt}
\begin{equation} \label{proto_sigma}
\small
    % {{\sigma_{lc}^t}^2} =  \frac{1}{n^t_{lc}} \sum_{li=1}^{n^t_{lc}}(\mu_{lc}^t - f_{p_l^t}(x_{lci}))^2 \: , x_{lci} \in \mathcal{T}_{lc}^t
    {{\sigma_{lc}^t}^2} =  \frac{1}{|\mathcal{T}^t_{lc}|} \sum_{li=1}^{|\mathcal{T}^t_{lc}|}(\mu_{lc}^t - f_{p_l^{r,t}}(x_{lci}))^2 \: , x_{lci} \in \mathcal{T}_{lc}^t
\end{equation}
% \vspace{-2pt}
 Due to non-i.i.d data between clients, FCIL satisfies $\mathcal{T}^{t}_{l} \subseteq (\mathcal{T}^{t} = \cup_{l=1}^{L} \mathcal{T}^{t}_{l})$ and $\forall l, l' \in[1..L], l\neq l' \Rightarrow (\mathcal{T}^{t}_{l} \neq \mathcal{T}^{t}_{l'}) \wedge (\mathcal{C}^{t}_{l} \neq \mathcal{C}^{t}_{l'})$. It implies a client only optimize $p_l$ and $\phi_l$ with regard to $\mathcal{T}_l^{t}$ but not yet to $\mathcal{T}^{t}$. The client $S_l$ only learns $C^t_l$ but not $\overline{\mathcal{C}^t_l}=\mathcal{C}^t-\mathcal{C}^t_l$, classes that exist in $\mathcal{T}^t$ but not in $\mathcal{T}^t_l$. Meanwhile, we can't afford to share any exemplar sample $(x^t_{li},y^t_{li})$ due to data privacy. Aggregating the prompts and head doesn't guarantee optimal global model $(p_G^{r,t}, \phi_G^{r,t})$ for $\mathcal{T}^{t}$ since in the next round, the participating clients and their available data may be different from the current rounds' $\forall r, r'\in[1..R], \forall l\in[1..L], r \neq r' \Rightarrow (S_l^{r,t} \neq S_l^{r',t}) \wedge ({\mathcal{T}}_l^{r,t} \neq {\mathcal{T}}_l^{r',t})$. Furthermore, training in too many rounds leads to catastrophic forgetting. 
 
 To handle this challenge, we propose prototype sharing between clients and the central server, where the server collects all clients' prototype $\{{Z}_l^{t}\}^L_{l=1}$ then distributes global prototype ${Z}^{t}_{G} = {Z}^{t} = \cup_{l=1}^{L} {Z}^{t}_{l}$. The shared prototype set $Z_l^t$ ensures each client learns $\overline{\mathcal{C}^t_l}$ via $\overline{Z^t_l}=Z^t-Z^t_l$, therefore each client has chances to learn all classes $\mathcal{C}^t$ in $\mathcal{T}^t$. Besides, the prototype set $Z^t$ has a small size in comparison to prompt $p_l$ or head layer $\phi_l$ where $z_c^t \in Z^t$ has the size of $1\times D$, where D is the embedding dimension, and It doesn't contain any raw information as contained in an exemplar sample.

 Now each client can enhance its local training by \textbf{injecting} the shareable prototype set $Z^t$ into its training process by considering $\mathcal{T}_l^{t} \cup (Z^t, \mathcal{C}^t) \approx \mathcal{T}^{t}$, where $Z^t$ is the prototype feature set, and $\mathcal{C}^t$ acts as the label for the prototypes of all classes $c \in \mathcal{C}^t$. Therefore, the loss function and parameters update of each client can be defined as in Eq. \ref{loss_client2} and \ref{param_update2}. In case the prototype is not yet available e.g. first round ($r=1$), a local model is updated using Eq. \ref{loss_client} and \ref{param_update}. 
 \begin{equation}\label{loss_client2}
 \small
 \begin{split}
     \mathcal{L}_{l+} = &\mathcal{L}(f_{\phi_{l}^{r,t}}(f_{p_l^{r,t}}(x_{li}^{t})\cup Z^t),y_{li}^{t}\cup \mathcal{C}^t)) \\
     &+ \lambda \mathcal{L}_{m}(x_l^{t},k_l^{r,t}), (x_{li}^{t},y_{li}^{t})\in \mathcal{T}_l^{t}
\end{split}
\end{equation}
\begin{equation}\label{param_update2}
\small
\begin{split}
    &p_l^{r,t} \gets p_l^{r,t} - \alpha \nabla \mathcal{L}_l+\\
    &\phi_l^{r,t} \gets \phi_l^{r,t} - \alpha \nabla \mathcal{L}_l+ 
\end{split}
\end{equation}
% \vspace{-15pt}
\subsection{Handling Class Imbalance via Prototype Augmentation}\label{prototype_augmentaion}
% \vspace{-5pt}
We can simply assign $z_{c}^t \in Z^t$ by $\mu_{c}^t$ to generate a single prototype for class-c. However, to handle classes imbalance between locally available classes and unavailable classes that are represented by the shared prototypes.  we enrich the prototypes by using an augmentation.  Suppose that $x_{c_1}$ is the sample for class $c_1 \in {\mathcal{C}^t_l}$, $z_{c_2}^t \in Z^t$ is the prototype of class $c_2 \in \overline{\mathcal{C}^t_l} = \mathcal{C}^t - \mathcal{C}_l^t$, $c_1$ is locally available class, and $c_2$ is locally the unavailable class in client-$l$, then we have $|x_{c_1}| \gg 1$, while $|z_{c_2}^t|= 1$. The prototype augmentations create artificial prototypes that satisfy  $|z_{c_2}^t| \approx |x_{c_1}|$ as defined in equation \ref{proto_aug} to generate $m$ augmented prototypes for class-$c$ based on Gaussian distribution $z_{c}^t \sim \mathcal{N}(\mu_{c}^t,\,{\sigma_{c}^t}^{2})$, and $\beta$ is a random value in $(0,1)$ range.
\begin{equation} \label{proto_aug}
\small
\begin{split}
        &z_{c}^t = \{\mu_{c}^t\} \cup \{u_{ci}^t\}_{i=1}^{m},\:where\: \\ 
        &u_{ci}^t=\mu_{c}^t+\beta{\sigma_{c}^t}, m \geq 1
\end{split}
\end{equation}

\subsection{Server Weighted Aggregation}\label{weighted_aggregation}
% \vspace{-5pt}
We proposed weighted Gaussian aggregation on the server side to improve global model generalization. The aggregation treats clients' contribution to global aggregation proportionally based on their participation and their training sample size following best practice where a model that learns more produces more convergent weight. We consider clients to learn their local data and then produce Gaussian distributed local parameters. We define $\omega_l=\rho_l^{t}|\mathcal{T}_l^{t}|$ as the weight of a client-l on the t-th task, where $\rho_t$ is the total of client-l participation until $t-th$ task and $|\mathcal{T}_l^{t}|$ is the number of samples in $\mathcal{T}_l^t$. Given $\{(p_l^{r,t},\phi_l^{r,t},Z_l^{t})\}_{l=1}^{L}$ is a set of locally optimized parameters by selected local clients $\{S_l\}_{l=1}^L$ on the $r-th$ of $t-th$ task, then the global parameters $(p_G^{r,t}$, $\phi_G^{r,t}$ and $Z_G^t$ are computed by Gaussian-based weighted aggregation as We define in Eq. \ref{wagg_prompt}-\ref{wagg_proto2}. Equation \ The derivation  of the proposed weighted aggregation in presented in our supplemental document.%Appendix \ref{app:wagg}%\textbf{Appendix A}.%
The pseudo-code of PIP is presented in algorithm \ref{alg:pip}, while the pseudocode of our proposed baselines is presented in our supplemental document. 
% \vspace{-12pt}
%Algorithm \ref{alg:pip} in Appendix \ref{app:pip}%  
\begin{equation}\label{wagg_prompt}
\small
    {p_{G}^{r,t}}= \frac{1}{\sum_{l=1}^{L}\omega_l}\sum_{l=1}^{L}(p^{r,t}_{l}\omega_{l})
\end{equation}
\vspace{-2pt}
\begin{equation}\label{wagg_head}
\small
    {\phi_{G}^{r,t}}= \frac{1}{\sum_{l=1}^{L}\omega_l}\sum_{l=1}^{L}(\phi^{r,t}_{l}\omega_{l})
\end{equation}
\vspace{-2pt}
\begin{equation} \label{wagg_proto}
\small
    \begin{split}
    &z_{Gc}^t=\mathcal{N}(\mu_{Gc}^t,{\sigma_{Gc}^t}^2),\: where \\[-3pt]
    &\mu_{Gc}^t = \frac{1}{\sum_{l=1}^{L}a_{lc}^t\omega_l}\sum_{l=1}^{L} a_{lc}^t \mu_{lc}^t\omega_{l}\\
    \end{split}
\end{equation}
\vspace{-2pt}
\begin{equation} \label{wagg_proto2}
\small
    \begin{split}
    &{\sigma_{Gc}^t}^2 = \frac{1}{\sum_{l=1}^{L}a_{lc}^t\omega_l}\sum_{l=1}^{L} a_{lc}^t ({\mu_{lc}^t}^2 + {\sigma_{lc}^t}^2)\omega_{l} - {\mu_{Gc}^t}^2\\
    &where\: a_{lc}^t = 1, \: if(\mu_{lc}^t\;exists)\: else\;=0
    \end{split}
\end{equation}

\begin{algorithm}
\caption{PIP}\label{alg:pip}
\small
\begin{algorithmic}[1]
% \Require $n \geq 0$
\State \textbf{Input:} Number of clients $N$, number of selected local clients $L$, total number of rounds $R$, number of task $T$, local epochs $E$,  batch size $B$.
% \KwInput{Number of clients $N$, number of selected local clients $L$, total number of rounds $R$, number of task $T$, local epochs $E$,  batch size $B$, number}
% \item[\hspace{3 mm} // Some label if needed]
\State Distribute frozen ViT backbone $f$ to all clients $\{S_l\}_{l=1}^{N}$ and central server $S_G$ 
\State Initiate prompt, key, and head layer for all clients and central server $p_G = p_l$, $\phi_G = \phi_l,\, l \in \{1..N\}$ 
\State $R_T \leftarrow R/T$, $R_T$ represents round per task
\For{$t=1:T$}
    \State Init global and local prototypes $Z_G^t=Z_l^t,=Z^t=\emptyset$
    \For{$r=1:R_T$}
        \State $S_l \leftarrow$ randomly select $L$ local clients from $N$ total clients
        \State \textbf{Clients execute}: 
        \State Receive global parameters i.e  ($p_G,\,Z_G^t$, $\phi_G$) 
        \State Assign local parameters $(p_l,\phi_l,Z_l^t) \leftarrow$  $(p_G,\phi_G,Z_G^t)$ 
        \State $\mathcal{B} \leftarrow$ Split $\mathcal{T}_l^t$ into $B$ sized batches
        \For{$e=1:E$}
            \For{$b=1:\mathcal{B}$}
                \State Compute feature $f_{p_l}(x)$ as in Eq. (1) to (4)
                \State Compute logits with prototypes $f_{\phi_l}(f_{p_l}(x) \cup Z^t)$
                \State Compute loss $\mathcal{L}_{t+}$ as in Eq. (9)
                \State Update local parameters $(p_l,\phi_l)$ w.r.t $\mathcal{L}_{t+}$ as in Eq. (10)  
            \EndFor
            \If{$Z_l^t=\emptyset$}
                % \State Compute prototype $z^t_{lc} \sim \mathcal{N}(\mu_{lc}^t,\,{\sigma_{lc}^t}^2)$ for all class-$c$ available in $\mathcal{T}^{t}_l$ as in Eq. (7) and (8)
                \State Compute prototype $z^t_{lc} \sim \mathcal{N}(\mu_{lc}^t,\,{\sigma_{lc}^t}^2)$ as in Eq. (7)-(8)
            \EndIf
            \State Augment and unify the prototypes as in Eq. (11) 
        \EndFor
        % \State Compute prototype $z^t_{lc} \sim \mathcal{N}(\mu_{lc}^t,\,{\sigma_{lc}^t}^2)$ for all class-$c$ available in $\mathcal{T}^{t}_l$ as in Eq. (7) and (8)
        \State Compute prototype $z^t_{lc} \sim \mathcal{N}(\mu_{lc}^t,\,{\sigma_{lc}^t}^2)$ as in Eq. (7)-(8)
        \State Set $Z_l^t=\{z_{lc}^t\}=\{(\mu_{lc}^t,\,{\sigma_{lc}^t}^2)\}$, for class-$c$ available in $\mathcal{T}^{t}_l$
        \State Store local parameters $(p_l,\phi_l,\,Z_l^t)$
        \State Compute clients' weight $\omega_l^t$
        \State Send local parameters $(p_l,\phi_l,\,Z_l^t)$ and weight $\omega_l^t$ to server
        \State \textbf{Server executes}:
        \State Receives selected local clients parameters $\{(p_l,\phi_l,\,Z_l^t)\}_{l=1}^{l=L}$ 
        \State Aggregates parameters $(p_G,\phi_G)\leftarrow(p_l,\phi_l)$ as in Eq. (12)-(13)
        \State Aggregates prototypes  $Z_G^t \leftarrow Z_l^t$ as in Eq. (14)-(15)
        \State Send global parameters  $(p_G,\phi_G,Z_G^t)$ to all clients   
    \EndFor
\EndFor
\State \textbf{Output:}  Optimal Global parameters $(p_G,\phi_G)$ and local parameters $(p_l,\phi_l), l \in {1..N}$  
\end{algorithmic}
\end{algorithm}

\section{Theoretical Analysis}
% \vspace{-5pt}
Let $\theta=(p,\phi)$ is the trainable parameters and $F(\theta)=\mathbb{E}[\mathcal{L}_{l+}(\mathcal{T};\theta)]=\mathbb{E}[\mathcal{L}_{l+}(\mathcal{T};(p,\phi))]$ is the expected loss function, $k,E$, $R$, and $L_S$ is local iteration, local epoch, global round, and number of selected local clients respectively. We follow $L$-smooth and $\mu$-strongly convex $F$, random uniformly distributed batches, $G$-bounded uniformly gradient assumptions, and decreasing learning rate as in \cite{li2019convergence},\cite{bottou2018optimization} as detailed below:

\noindent\textbf{Assumption 1:} $F_1,...F_l,...,F_{L_S}$ are all $L-$smooth: for all $\theta$ and $\theta'$, $F_l(\theta) \leq F_l(\theta') + (\theta - \theta')^T \nabla F_l(\theta) + \frac{L}{2} ||\theta-\theta'||_2^2.$

\noindent\textbf{Assumption 2:} $F_1,...F_l,...,F_{L_S}$  are all $\mu-$strongly convex: for all $\theta$ and $\theta'$, $F_l(\theta) \leq F_l(\theta') + (\theta - \theta')^T \nabla F_l(\theta) + \frac{\mu}{2} ||\theta-\theta'||_2^2.$

\noindent\textbf{Assumption 3:} Let $\xi_l^k$ be the random uniformly sampled from $l$-th local data at $k-th$ iteration . The variance of stochastic gradients in each client is bounded by: $\mathbb{E}||\nabla F_l(\theta_l^k, \xi_l^k) - \nabla F_l(\theta_l^k)|| \leq \sigma_l^2$ for $l=1,2,..., L_S$

\noindent\textbf{Assumption 4:}The expected squared norm of stochastic gradients in each client is bounded by: $\mathbb{E}||\nabla F_l(\theta_l^k, \xi_l^k)|| \leq G^2$ for all $l=1,2,...,L_S$ and $k=1,2,...., K$ where $K \in \mathbb{N}$.

\noindent\textbf{Assumption 5:} $\sum_{k=1}^{\infty} \alpha_l^k = \infty$ and $\sum_{k=1}^{\infty} {\alpha_l^k}^2 < \infty$ where  $\alpha_l^k$ is the learning rate of $l-th$ client in $k$-th step training.

\noindent\textbf{Assumption 6:} The objective function $F_l$ and SG satisfy the following conditions.

\noindent(a). The sequence of $\{\theta_l^k\}$ is contained in an open space where $F_l$ is bounded below by a scalar $F_{inf}$

\noindent(b) Exist scalars $\nu_G \geq \nu > 0$ so that for all $k \in \mathbb{N}$ satisfy:
\begin{equation} \label{}
\small
\begin{split}
     &\nabla F_l(\theta_l^k)^T \mathbb{E}_{\xi_l^k}[g(\theta_l^k,\xi_l^k)] \geq \nu||\nabla F_l(\theta_l^k)T||_2^2, and \\
     &|| \mathbb{E}_{\xi_l^k}[g(\theta_l^k,\xi_l^k)]||_2 \leq \nu_G||\nabla F_l(\theta_l^k)||_2.
\end{split}
\end{equation}

\noindent We state the following theorems:

\noindent \textbf{Theorem 1:} $\liminf_{k \rightarrow \infty} \mathbb{E}[||\nabla F(\theta_k)||_2^2]=0$

\noindent \textbf{Theorem 2:} Let Chose $\kappa=L/\mu$, $\delta=\max(8\kappa,E)$, $\alpha_r=2/(\mu(\delta+r))$, $C=4/L_SE^2G^2$, $\theta^1,\theta^R, \theta^*$ is the initial, last updated ($R$-th), and optimum parameter respectively, $F^*$ is minimum value of $F$ then: 

\noindent$\mathbb{E}[F(\theta^R)]-F^* \leq \frac{\kappa}{(\delta+R-1)}(\frac{2(B+C)}{\mu} +\frac{\mu\delta}{2}\mathbb{E}||\theta^1-\theta^*||)$. 

\noindent \textbf{Theorem 3:} Given $\theta^*$ and $\theta$ are optimal parameter in $\mathcal{T}^t_l \cup Z^t$ and $\mathcal{T}^t_l$ respectively, where $\mathcal{T}^{t}_{l} \subset \mathcal{T}^{t}$, where  $|\mathcal{T}^{t}_{l}| / |\mathcal{T}^{t}| = \eta \in (0,1)$, then at least there's $\epsilon > 0$ that satisfy $F(\theta;\mathcal{T}^t)- F(\theta^*;\mathcal{T}^t) \geq \epsilon$.
Theorem 1 and 2 prove PIP local training and federated convergence respectively, while theorem 3 proves PIP generalization. The detailed theoretical analysis, and derivations are presented in supplemental document.

\section{Experiment Results and Analysis}
% \vspace{-3pt}
\subsection{Experimental Setting}
% \vspace{-3pt}
\noindent\textbf{Datasets:} Our experiment is conducted using three main benchmarks in FCIL i.e. split CIFAR100, split MiniImageNet, and split TinyImageNet. The CIFAR100 and miniImageNet datasets each contain 100 classes while TinyImagenet is a dataset with 200 classes. We follow settings from \cite{dong2022federated} and \cite{dong2023no} where the dataset is split equally into all tasks. In our main numerical result, The dataset is split into 10 tasks i.e. 10 classes per task for CIFAR100 and MiniImageNet, and 20 classes per task for the TinyImageNet dataset. In our further analysis, we investigate the performance of the proposed methods in different task sizes e.g. T=5 and T=20.

\noindent\textbf{Benchmark Algorithms:} PIP is implemented in two version algorithms i.e. PIP-L2P and PIP-DualP with L2P and DualP prompt structure respectively, PIP-L2P and PIP-DualP are compared with 10 state-of-the-art algorithms: LGA \cite{dong2023no}, TARGET \cite{zhang2023target}, GLFC\cite{dong2022federated}, AFC\cite{kang2022class}+FL, DyTox\cite{douillard2022dytox}+FL, SS-IL\cite{ahn2021ss}+FL, GeoDL\cite{simon2021learning}+iCaRL\cite{rebuffi2017icarl}+FL, DDE\cite{hu2021distilling}+iCaRL\cite{rebuffi2017icarl}+FL, PODNet\cite{douillard2020podnet}+FL, BiC\cite{wu2019large}+FL, iCaRL\cite{rebuffi2017icarl}+FL, and the proposed baseline models (FedL2P and Fed-DualP.). We compare with Fed-CPrompt\cite{bagwe2023fed} in the CIFAR100 dataset since it reported evaluation in the CIFAR100 dataset only. We didn't include a performance report from FCILPT\cite{liu2023federated} since they use memory in their experiment which is different from our setting. Please note that we can't find the complete and runnable released code from both Fed-CPrompt and FCILPT to produce the results in our simulation.

\noindent\textbf{Experimental Details:} Our numerical study is executed under a single NVIDIA A100 GPU with 40 GB memory across 3 runs with different random seeds \{2021,2022,2023\}. Fed-L2P, Fed-DualP, PIP-L2P, and  PIP-DualP train $T$ number of prompts $p \in R ^{5\times 768}$ and $\phi \in R^{|\mathcal{C}\times 768|}$ head layer, while the competitors train whole CNN models following their original implementation. Following \cite{dong2023no} and \cite{dong2022federated}, each experiment is simulated by 30 total clients and 1 global server, where in each round, 10 (33.33\%) local clients are selected randomly. Each client randomly receives 60\% ($\eta=0.6$) class label space.  The total global round is set to 100, the local clients' epoch is set to 2, and the learning rate is set by choosing the best value from \{0.02,0.002\}. On each task, we evaluate average accuracy and forgetting from all learned tasks. 

\begin{table}[h]
\centering
\footnotesize
% \small
\setlength{\tabcolsep}{0.13em}
\begin{tabular}{l|ccccccccccccc}
\hline
                          & \multicolumn{13}{c}{Accuracy in each task (T) on CIFAR100 (T=10, @10classes/task)}                                                                                                                                                            \\ \hline
Method                    & 1            & 2            & 3            & 4            & 4           & 6           & 7           & 8           & 9           & 10           & Avg           & PD           & Imp  \\ \hline
iCaRL+FL                  & 89.0          & 55.0          & 57.0          & 52.3          & 50.3          & 49.3          & 46.3          & 41.7          & 40.3          & 36.7          & 51.8          & 52.3          & 36.2       \\
BiC+FL                    & 88.7          & 63.3          & 61.3          & 56.7          & 53.0          & 51.7          & 48.0          & 44.0          & 42.7          & 40.7          & 55.0          & 48.0          & 33.0       \\
PODNet+FL                 & 89.0          & 71.3          & 69.0          & 63.3          & 59.0          & 55.3          & 50.7          & 48.7          & 45.3          & 45.0          & 59.7          & 44.0          & 28.3       \\
DDE+iCaRL+FL              & 88.0          & 70.0          & 67.3          & 62.0          & 57.3          & 54.7          & 50.3          & 48.3          & 45.7          & 44.3          & 58.8          & 43.7          & 29.2       \\
GeoDL+iCaRL+FL            & 87.0          & 76.0          & 70.3          & 64.3          & 60.7          & 57.3          & 54.7          & 50.3          & 48.3          & 46.3          & 61.5          & 40.7          & 26.5       \\
SS-IL+FL                  & 88.3          & 66.3          & 54.0          & 54.0          & 44.7          & 54.7          & 50.0          & 47.7          & 45.3          & 44.0          & 54.9          & 44.3          & 33.1       \\
DyTox+FL                  & 86.2          & 76.9          & 73.3          & 69.5          & 62.1          & 62.7          & 58.1          & 57.2          & 55.4          & 52.1          & 65.4          & 34.1          & 22.7       \\
AFC+FL                    & 85.6          & 73.0          & 65.1          & 62.4          & 54.0          & 53.1          & 51.9          & 47.0          & 46.1          & 43.6          & 58.2          & 42.0          & 29.8       \\
GLFC                      & 90.0          & 82.3          & 77.0          & 72.3          & 65.0          & 66.3          & 59.7          & 56.3          & 50.3          & 50.0          & 66.9          & 40.0          & 21.1       \\
LGA                       & 89.6          & 83.2          & 79.3          & 76.1          & 72.9          & 71.7          & 68.4          & 65.7          & 64.7          & 62.9          & 73.5          & 26.7          & 14.6       \\
TARGET                    & 70.6          & 39.0          & 27.5          & 21.6          & 18.8          & 16.4          & 14.2          & 12.2          & 10.7          & 9.1           & 24.0          & 61.4          & 64.0       \\
Fed-CPrompt               & 97.3          & 90.5         & 86.7         & 83.4         & 83.3          & 82.2         & 81.6         & 79.7          & 79.0         & 79.4         & 84.3          & 17.9          & 3.7        \\
Fed-L2P (BL)              & 94.4          & 73.9          & 68.6          & 67.5          & 64.7          & 64.8          & 66.6          & 66.6          & 66.2          & 65.9          & 69.9          & 28.6          & 18.1       \\
Fed-DualP (BL)            & 96.8          & 81.6          & 75.7          & 72.6          & 69.5          & 67.3          & 69.9          & 69.2          & 69.9          & 69.7          & 74.2          & 27.1          & 13.8       \\
\textbf{PIP-L2P (Ours)}   & 95.4          & 90.1          & 86.7          & 84.7          & 82.9          & 80.9          & 80.1          & 79.5          & 78.5          & 77.5          & 83.6          & 17.9          & 4.4        \\
\textbf{PIP-DualP (Ours)} & \textbf{98.7} & \textbf{92.9} & \textbf{89.4} & \textbf{87.6} & \textbf{87.0} & \textbf{85.3} & \textbf{85.2} & \textbf{84.8} & \textbf{84.8} & \textbf{84.3} & \textbf{88.0} & \textbf{14.4} & \textbf{-}
\\ \hline
\end{tabular}
\caption{\footnotesize Numerical results on CIFAR100 (T=10), "BL" denotes our proposed baseline, "Avg" denotes the average accuracy of all tasks, denotes performance drop, and "Imp" denotes improvement/gap of PIP-DualP compared to the respective method}
\label{tab:main_1dataset}
\end{table}

\begin{table*}[h!]
\centering
% \scriptsize
\footnotesize
\setlength{\tabcolsep}{0.3em}

\begin{tabular}{l|ccccccccccccc|ccccccccccccc}
\hline
                          & \multicolumn{13}{l}{Accuracy in each task (T) on MiniImageNet (T=10, @10classes/task)}                                                                                                              & \multicolumn{13}{l}{Accuracy in each task (T) on TinyImageNet (T=10, @20classes/task)}                                                                                                              \\ \hline
Method                    & 1             & 2             & 3             & 4             & 5             & 6             & 7             & 8             & 9             & 10            & Avg           & PD           & Imp  & 1             & 2             & 3             & 4             & 5             & 6             & 7             & 8             & 9             & 10            & Avg           & PD           & Imp  \ \\ \hline
iCaRL+FL                  & 74.0          & 62.3          & 56.3          & 47.7          & 46.0          & 40.3          & 37.7          & 34.3          & 33.3          & 32.7          & 46.5          & 41.3         & 44.6 & 63.0          & 53.0          & 48.0          & 41.7          & 38.0          & 36.0          & 33.3          & 30.7          & 29.7          & 28.0          & 40.1          & 35.0         & 46.1 \\
BiC+FL                    & 74.3          & 63.0          & 57.7          & 51.3          & 48.3          & 46.0          & 42.7          & 37.7          & 35.3          & 34.0          & 49.0          & 40.3         & 42.0 & 65.3          & 52.7          & 49.3          & 46.0          & 40.3          & 38.3          & 35.7          & 33.0          & 31.7          & 29.0          & 42.1          & 36.3         & 44.1 \\
PODNet+FL                 & 74.3          & 64.0          & 59.0          & 56.7          & 52.7          & 50.3          & 47.0          & 43.3          & 40.0          & 38.3          & 52.6          & 36.0         & 38.5 & 66.7          & 53.3          & 50.0          & 47.3          & 43.7          & 42.7          & 40.0          & 37.3          & 33.7          & 31.3          & 44.6          & 35.4         & 41.7 \\
DDE+iCaRL+FL              & 76.0          & 57.7          & 58.0          & 56.3          & 53.3          & 50.7          & 47.3          & 44.0          & 40.7          & 39.0          & 52.3          & 37.0         & 38.7 & 69.0          & 52.0          & 50.7          & 47.0          & 43.3          & 42.0          & 39.3          & 37.0          & 33.0          & 31.3          & 44.5          & 37.7         & 41.8 \\
GeoDL+iCaRL+FL            & 74.0          & 63.3          & 54.7          & 53.3          & 50.7          & 46.7          & 41.3          & 39.7          & 38.3          & 37.0          & 49.9          & 37.0         & 41.1 & 66.3          & 54.3          & 52.0          & 48.7          & 45.0          & 42.0          & 39.3          & 36.0          & 32.7          & 30.0          & 44.6          & 36.3         & 41.6 \\
SS-IL+FL                  & 69.7          & 60.0          & 50.3          & 45.7          & 41.7          & 44.3          & 39.0          & 38.3          & 38.0          & 37.3          & 46.4          & 32.4         & 44.6 & 62.0          & 48.7          & 40.0          & 38.0          & 37.0          & 35.0          & 32.3          & 30.3          & 28.7          & 27.0          & 37.9          & 35.0         & 48.4 \\
DyTox+FL                  & 76.3          & 68.3          & 64.8          & 58.6          & 45.4          & 41.3          & 39.7          & 37.1          & 36.2          & 35.3          & 50.3          & 41.0         & 40.7 & 73.2          & 66.6          & 48.0          & 47.1          & 41.6          & 40.8          & 37.4          & 36.2          & 32.8          & 30.6          & 45.4          & 42.6         & 40.8 \\
AFC+FL                    & 82.5          & 74.1          & 66.8          & 60.0          & 48.0          & 44.3          & 42.5          & 40.9          & 39.0          & 36.1          & 53.4          & 46.4         & 37.6 & 73.7          & 59.1          & 50.8          & 43.1          & 37.0          & 35.2          & 32.6          & 32.0          & 28.9          & 27.1          & 42.0          & 46.6         & 44.3 \\
GLFC                      & 73.0          & 69.3          & 68.0          & 61.0          & 58.3          & 54.0          & 51.3          & 48.0          & 44.3          & 42.7          & 57.0          & 30.3         & 34.0 & 66.0          & 58.3          & 55.3          & 51.0          & 47.7          & 45.3          & 43.0          & 40.0          & 37.3          & 35.0          & 47.9          & 31.0         & 38.4 \\
LGA                       & 83.0          & 74.2          & 72.3          & 72.2          & 68.1          & 65.8          & 64.0          & 59.6          & 58.4          & 57.5          & 67.5          & 25.5         & 23.5 & 70.3          & 64.0          & 60.3          & 58.0          & 55.8          & 53.1          & 47.9          & 45.3          & 39.8          & 37.3          & 53.2          & 33.0         & 33.1 \\
TARGET                    & 56.1          & 27.5          & 20.0          & 15.9          & 11.9          & 8.8           & 6.9           & 5.0           & 2.7           & 2.3           & 15.7          & 53.8         & 75.3 & 57.1          & 32.7          & 20.9          & 14.7          & 11.6          & 8.3           & 7.1           & 5.2           & 3.8           & 3.0           & 16.4          & 54.1         & 69.8 \\
Fed-L2P (BL)              & 95.2          & 80.9          & 79.1          & 77.7          & 76.3          & 75.7          & 76.0          & 75.8          & 74.1          & 76.7          & 78.7          & 18.6         & 12.3 & 80.6          & 71.5          & 70.6          & 67.5          & 65.2          & 66.7          & 66.2          & 65.9          & 64.2          & 64.3          & 68.3          & 16.3         & 18.0 \\
Fed-DualP (BL)            & 97.6          & 83.6          & 82.5          & 79.2          & 76.6          & 76.2          & 75.8          & 75.8          & 75.6          & 78.5          & 80.1          & 19.1         & 10.9 & 86.3          & 74.6          & 71.2          & 65.9          & 63.3          & 62.0          & 61.3          & 59.9          & 58.2          & 61.3          & 66.4          & 25.0         & 19.9 \\
\textbf{PIP-L2P (Ours)}   & 96.1          & 91.9          & 89.6          & 88.9          & 88.2          & 88.2          & 87.4          & 87.2          & 85.5          & 85.7          & 88.9          & 10.5         & 2.2  & 85.4          & 84.0          & 83.5          & 84.5          & 83.3          & 83.7          & 82.3          & 81.3          & 79.8          & 77.4          & 82.5          & \textbf{7.9} & 3.7  \\
\textbf{PIP-DualP (Ours)} & \textbf{98.4} & \textbf{90.9} & \textbf{90.4} & \textbf{90.7} & \textbf{90.2} & \textbf{90.7} & \textbf{90.3} & \textbf{90.4} & \textbf{88.9} & \textbf{89.2} & \textbf{91.0} & \textbf{9.2} & -    & \textbf{92.8} & \textbf{86.4} & \textbf{86.6} & \textbf{87.5} & \textbf{86.7} & \textbf{87.0} & \textbf{85.3} & \textbf{84.9} & \textbf{83.9} & \textbf{81.4} & \textbf{86.3} & 11.4         & -      \\ \hline
\end{tabular}
\caption{ Numerical results on MniImageNet and TinyImageNet (T=10), "BL" denotes our proposed baseline, "Avg" denotes the average accuracy of all tasks, "PD" denotes performance drop, and "Imp" denotes improvement/gap of PIP-DualP compared to the respective method}
\label{tab:main_2dataset}
\end{table*}

\begin{table}[]
\centering
% \scriptsize
\small
\setlength{\tabcolsep}{0.3em}
\begin{tabular}{lccccccccccc}
\hline
\multirow{2}{*}{Method} & \multicolumn{10}{c}{Average Forgetting on each Task}           & \multirow{2}{*}{AVG} \\ \cline{2-11}
                        & 1 & 2    & 3   & 4   & 5    & 6    & 7    & 8    & 9    & 10   &                      \\ \hline
LGA                     & - & 4.0  & 4.4 & 8.9 & 13.7 & 12.6 & 13.9 & 14.7 & 15.7 & 16.8 & 11.6                 \\
TARGET                  & - & -0.6 & 0.4 & 1.7 & 1.7  & 2.0  & 2.5  & 3.8  & 4.7  & 4.7  & 2.3                  \\
Fed-CPrompt             & - & 3.8  & 5.0 & 5.3 & 5.2  & 5.2  & 5.1  & 5.4  & 5.3  & 4.8  & 5.0                  \\
Fed-L2P                 & - & 0.9  & 1.9 & 0.3 & -0.4 & -3.4 & -5.0 & -4.0 & -3.2 & -4.0 & -1.9                 \\
Fed-DualP               & - & -0.4 & 2.7 & 2.3 & 2.4  & 2.8  & -0.9 & 0.6  & 0.4  & 0.6  & 1.2                  \\
PIP-L2P                 & - & 2.3  & 4.7 & 5.1 & 6.0  & 6.5  & 6.0  & 6.4  & 6.4  & 8.4  & 5.8                  \\
PIP-DualP               & - & 1.3  & 1.5 & 1.9 & 2.3  & 3.1  & 2.9  & 2.4  & 2.5  & 2.8  & 2.3 \\ \hline                
\end{tabular}
\caption{Detailed average forgetting of the consolidated methods in CIFAR100 (T=10)}
\label{tab:forgetting}
\end{table}

\begin{table}[h]
% \vspace{-9pt}
% \scriptsize
\small
\setlength{\tabcolsep}{0.2em}
\centering
\begin{tabular}{cccccccccc}
\hline
Conf.      & Prompt & \begin{tabular}[c]{@{}c@{}}Proto\\+Aug\end{tabular} & W.Agg & Head & \begin{tabular}[c]{@{}c@{}}T1\\(C=10)\end{tabular}      & \begin{tabular}[c]{@{}c@{}}T10\\(C=100)\end{tabular}    & Avg            & PD             & Imp        \\ \hline
% Config.      & Prompt & Proto+Aug & W.Agg & Head & T1 (C=10)      & T10 (C=100)    & Avg            & PD             & Imp        \\ \hline
A            & {\checkmark}      &       &       &      & 82.15          & 60.66          & 65.69          & 21.49          & 22.59      \\
B            & {\checkmark}      & {\checkmark}     &       &      & 90.99          & 77.56          & {81.59} & \textbf{13.43} & 6.69       \\
C            & {\checkmark}      &       & {\checkmark}     &      & 82.13          & 61.35          & 66.10          & 20.78          & 22.18      \\
D            & {\checkmark}      & {\checkmark}     & {\checkmark}     &      & 91.37          & 78.47          & {82.41} & \textbf{12.90} & 5.87       \\
E            & {\checkmark}      &       &       & {\checkmark}    & 96.60          & 70.68          & 75.38          & 25.92          & {12.90}      \\
F            & {\checkmark}      & {\checkmark}     &       & {\checkmark}    & \textbf{98.90}              & 84.17              & 87.82               & 14.73              & 0.46          \\
G            & {\checkmark}      &       & {\checkmark}     & {\checkmark}    & 98.60          & 69.94          & 72.57          & 28.66          & 15.71      \\ 
\textbf{PIP} & {\checkmark}      & {\checkmark}     & {\checkmark}     & {\checkmark}    & \textbf{98.60} & \textbf{84.60} & \textbf{88.28} & \textbf{14.00} & \textbf{-} \\ \hline
\end{tabular}
\caption{Ablation study on CIFAR100 dataset in one-seeded run i.e. 2021 using DualP prompt structure.  "Avg" denotes the average accuracy of all tasks, "PD" denotes performance drop, and "Imp" denotes improvement/gap of PIP compared to the respective configuration}
\label{tab:ablation}
\end{table}

\subsection{Numerical Results}
% \vspace{-5pt}
The numerical result of the consolidated algorithms is shown in tables \ref{tab:main_1dataset} and \ref{tab:main_2dataset}.  Except to Fed-CPrompt, The baseline method (Fed-DualP) already achieves higher average accuracy (Avg) than the SOTA methods with $1-13\%$ improvement in accuracy. Fed-DualP also experiences a lower performance drop (PD) ($19-27\%$) compared to the SOTA methods ($\geq26\%$) except in the CIFAR100 dataset vs. LGA. Fed-L2P achieves higher performance than SOTAs in MiniImageNet and TinyImageNet datasets with $\geq11\%$ gap, but lower performance in the CIFAR100 dataset. The proposed method (PIP-DualP) achieves the highest accuracy with $\geq 10\%$ gap compared to the baseline method, and $\geq 14\%$ gap compared to the competitor methods (except Fed-CPrompt). The proposed method also achieves the lowest performance drop with ($10-13\%$) gap compared to the baseline and $\geq 12\%$ gap compared to the competitor methods. The table shows that PIP achieves a higher gap in TinyImageNet which is a relatively more complex problem than in MiniImageNet and CIFAR100. Compared to Fed-CPrompt, our proposed method achieve a higher accuracy and lower performance drop with a significant margin i.e. $\geq 3\%$ 

Looking at performance on each task, tables \ref{tab:main_1dataset}, and \ref{tab:main_2dataset} show that PIP-DualP achieves the highest accuracy in all tasks in those three datasets. In the first task, the proposed method achieves higher performance with a small gap $1-4\%$ gap compared to the baseline method. However, with the increasing number of tasks, the gap gets higher e.g. $5-12\%$ in task-2, $5-20\%$ in task-3, and   $6-20\%$ in task-10 (last task).  It shows that the proposed method handles catastrophic forgetting better than the baseline method. PIP-L2P achieves second-best performance in all tasks except in the first task of the CIFAR100 dataset.

\subsection{Forgetting Analysis}
Table \ref{tab:forgetting} shows the average forgetting of the consolidated methods in each task T. The table shows that our proposed method (PIP-DualP) achieves smaller average forgetting than existing SOTAs i.e. LGA and Fed-CPrompt with a significant gap i.e. $2.7\%$ and $9.3\%$ margin respectively. Fed-DualP achieves a comparable average forgetting with TARGET, but looking at the trend, TARGET forgetting is increasing along with the number of tasks, while  PIP-DualP forgetting is relatively stable. Please note that TARGET achieves lower average accuracy than PIP-DualP with a $64\%$ gap. Our baselines i.e. Fed-DualP and Fed-L2P archives have lower average forgetting than PIP-DualP despite achieving lower accuracy than PIP-DualP with a huge gap i.e. ($>13\%$). This shows that the baselines achieve better stability (old task accuracy) but in exchange struggle to achieve plasticity (new task accuracy).

% \vspace{-5pt}
\subsection{Ablation Study}
% \vspace{-3pt}
We conducted an ablation study to investigate the contribution of each component of the proposed method. The result is summarized in Table \ref{tab:ablation}, while the detailed result is presented in our supplemental document. 
% \textbf{Appendix D}. %\ref{app:ablation}. 
The result shows that the prototype and augmentation contribute the most to the improvement of the performance as shown by the performance difference of configurations E vs. F ($12\%$), and G vs. H ($16\%$). The weighted aggregation improves performance up to 1\% as shown by the performance difference of PIP and configuration F. The head layer aggregation also plays an important role in the proposed method as shown that the absence of this component decreases performance with $\geq6\%$ margin. The absence of two components e.g. prototype and head aggregation (configuration C) or prototype and weighted aggregation (configuration E) impact significantly the performance with 22\% and 13\% accuracy drop respectively. The absence of three components i.e. prototype, weighted aggregation, and head aggregation (configuration A) cost a 23\% performance drop in the method.

% \vspace{-7pt}
\subsection{Further Analayis}
\begin{figure}[h]
% \vspace{-5pt}
% \setlength{\abovecaptionskip}{-5pt plus 0pt minus 0pt}
% \setlength{\belowcaptionskip}{-20pt plus 0pt minus 0pt}
\begin{center}
%\framebox[4.0in]{$\;$}
\includegraphics[width=0.47\textwidth]{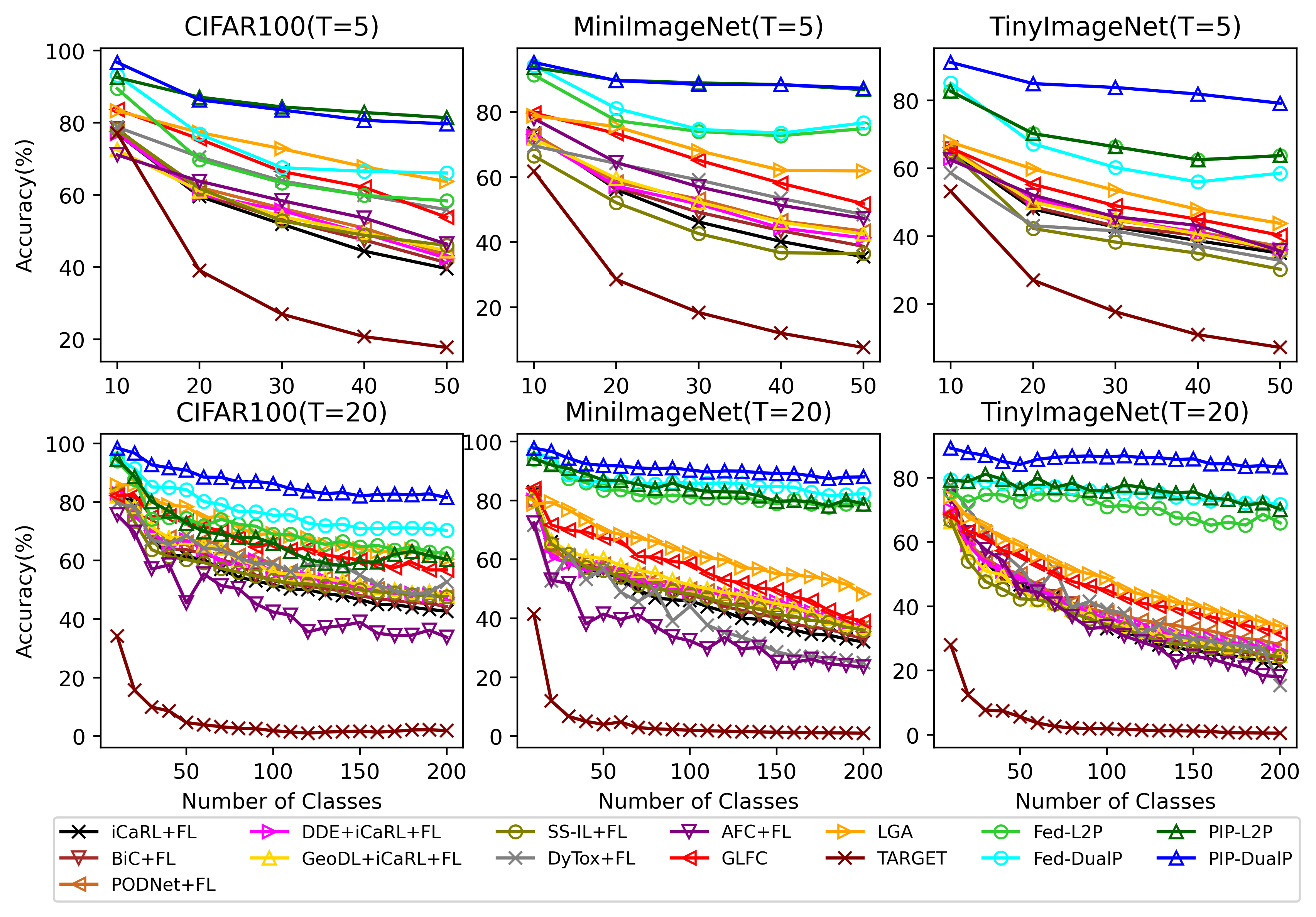}
\end{center}
\caption{Performance of the consolidated methods in CIFAR100, MiniImageNet and TinyImageNet with T=5 and T=20.}
\label{fig:task_size}
\end{figure}

% \vspace{-5pt}
\textbf{a) Different task size:} We evaluate the performance of the proposed method compared to the competitor methods in different task sizes i.e. T=5 and T=20 to further investigate the robustness of the proposed method. Figure \ref{fig:task_size} summarizes the performance of the consolidated methods in CIFAR100, MiniImageNet, and TinyImageNet with T=5 (upper figures) and T=20 (bottom figures). The detailed numerical result is presented in our supplemental document.. % \textbf{Appendix E}.
Both in T=5 and T=20 settings, PIP-DualP achieves the highest performance almost in every task in all configurations. It achieves a slightly lower performance than PiP-L2P in the last 3 tasks of CIFAR100 with T=5.  Besides, all figures show that the proposed method has gentle slopes compared to the baseline and competitor methods. It shows that the proposed method experiences the lowest performance drop from $t_{th}$ to $t+1_{th}$ task. It confirmed the robustness of the proposed method in different task-size settings. The baseline (Fed-DualP) method achieves better accuracy than the competitor methods in all 6 settings, except in CIFAR100 with the T=5 setting. In CIFAR100 with the T=5 setting, the baseline method achieves higher performance in task-1 and task-5, but lower performance in task-3, and comparable performance in task-2 and task-4. It shows the promising idea of the federated prompt-based approach for the FCIL problem.

% \vspace{-25pt}
\textbf{b) Small local clients:} We evaluate the performance of the proposed method in smaller selected local clients to advance our investigation of the robustness of the proposed method on smaller selected local clients. This simulates cross-device federated learning where in a round only a small portion of registered clients participate in the federated learning process. In our experiment with 30 total clients, we investigate the performance of the consolidated methods with 3 (10\%) and 2 (6.67\%) selected clients compared with the default setting with 10 clients (33.33\%). Table \ref{tab:smaller_clients} shows the performance of the consolidated methods in smaller local clients scenarios, while the complete result is presented in our supplemental document. 
% \textbf{Appendix F}. %\ref{app:smaller_nclients}. 
The figure shows that the proposed methods experience the lowest performance degradation in fewer local clients. The figures show that both in CIFAR100 and TinyImageNet datasets, the proposed method still achieves higher accuracy than the baseline and the competitor methods with 10 participating local clients.
% \vspace{-2pt}

\begin{figure}[h]
\begin{center}
%\framebox[4.0in]{$\;$}
\includegraphics[width=0.42\textwidth]{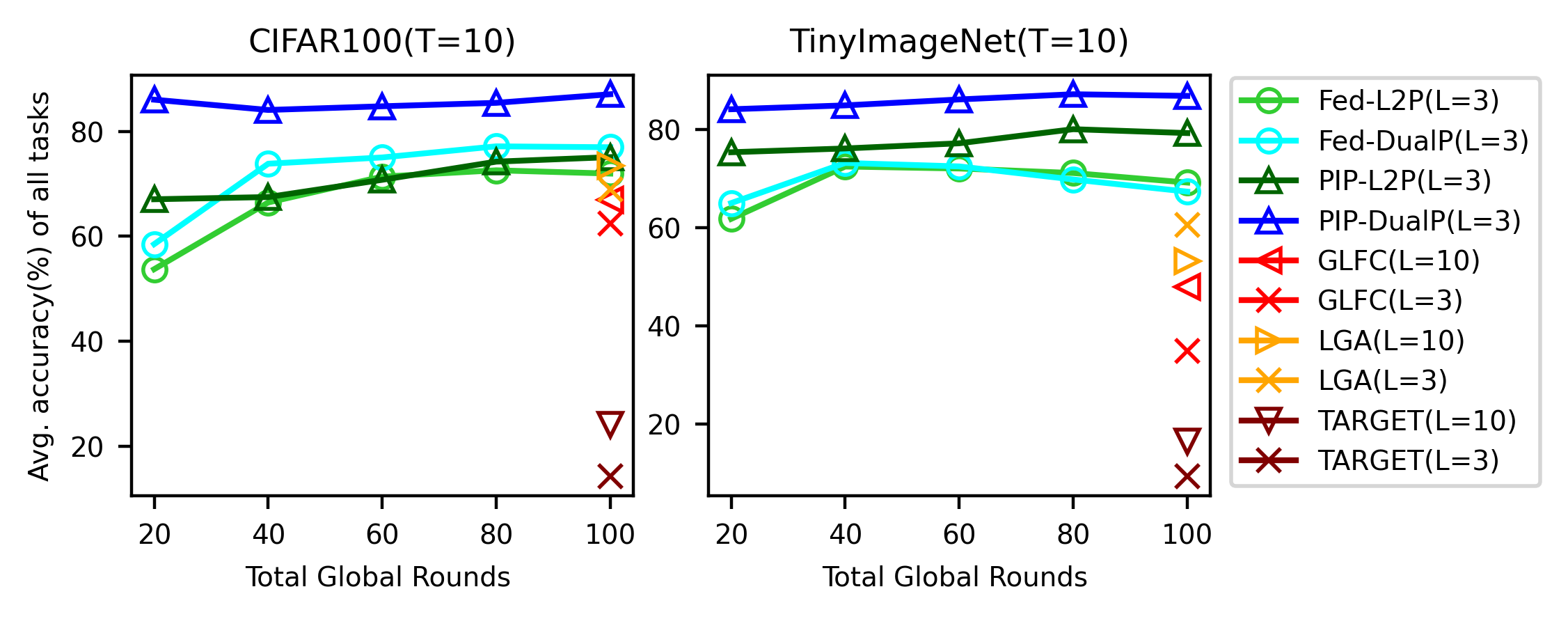}
\end{center}
\caption{Performance of the consolidated methods in CIFAR100 and TinyImageNet with smaller local clients (L) and smaller rounds.}
\label{fig:smaller_rounds}
\end{figure}

\textbf{c) Small global rounds:} We continue the previous experiment in smaller selected local clients into smaller local clients with smaller global rounds to study the robustness of the proposed method on a more extreme condition. This scenario simulates a condition where the model is urgently needed by the clients. In our experiment, we use 30 total clients and 3 (10\%) local clients. Figure \ref{fig:smaller_rounds} summarizes our investigation on the performance of the proposed methods with 20 to 100 (default) global rounds in CIFAR100 and TinyImageNet datasets, while the complete result is presented in our supplemental document. % \textbf{Appendix G}. 
The figure shows a common trend that the performance of the method is decreasing with the decreasing of global rounds.  However, our proposed methods (PIP-DualP and PIP-L2P) experience a relatively small amount of performance drop with the decreasing global rounds. Meanwhile, the baseline method experiences a significant drop in the CIFAR100 dataset when the number of global rounds is reduced to 20. Besides, It may experience over-fitting when the number of global rounds is too high as shown by the result on the TinyImageNet dataset. Compared to the current SOTA (GLFC and LGA) methods with the default number of global rounds (100), PIP-DualP still achieves higher performance, even though the number of global rounds is reduced to 20 (20\% default setting). 
\begin{table}[]
\centering
\small
% \scriptsize
% \setlength{\abovecaptionskip}{1pt plus 0pt minus 0pt}
% \setlength{\belowcaptionskip}{-10pt plus 0pt minus 0pt}
\setlength{\tabcolsep}{0.45em}
\begin{tabular}{lcccccc}
\hline
               & \multicolumn{3}{c}{CIFAR100 (T=10)}                                                                                                                                     & \multicolumn{3}{c}{TinyImageNet (T=10)}                                                                                                                                 \\ \hline
Method         & \begin{tabular}[c]{@{}c@{}}L=10\\ (33.3\%)\end{tabular} & \begin{tabular}[c]{@{}c@{}}L=3\\ (10\%)\end{tabular} & \begin{tabular}[c]{@{}c@{}}L=2\\ (6.67\%)\end{tabular} & \begin{tabular}[c]{@{}c@{}}L=10\\ (33.3\%)\end{tabular} & \begin{tabular}[c]{@{}c@{}}L=3\\ (10\%)\end{tabular} & \begin{tabular}[c]{@{}c@{}}L=2\\ (6.67\%)\end{tabular} \\ \hline
GLFC           & 66.92                                                   & 62.42                                                & 57.39                                                  & 47.89                                                   & 34.89$^o$                             & 32.77$^o$                                \\
LGA            & 73.45                                                   & 68.87                                                & 66.68                                                  & 53.18                                                   & 60.61$^*$                                               & 58.54$^*$                                                 \\
TARGET            & 24.00                                                   & 14.19                                                & 14.18                                                  & 16.43                                                   & 9.34                                               & 7.56                                                 \\ \hline
Fed-L2P        & 70.86                                                   & 71.94                                                & 71.09                                                  & 70.02                                                   & 69.16                                                & 65.25                                                  \\
Fed-DualP & 83.10                                                    & 75.08                                                & 64.68                                                  & 67.31                                                   & 66.11                                                & 67.03                                                  \\ 
% PIP-L2P        & 70.02                                                   & 69.16                                                & 65.25                                                  & 82.85                                                   & 79.23                                                & 76.17                                                  \\
% PIP-DualP & 82.85                                                   & 79.23                                                & 76.17                                                  & 86.79                                                   & 85.00                                                   & 82.13 
PIP-L2P        & 75.38                                                   & 76.96                                                & 72.17                                                  & 82.85                                                   & 79.23                                                & 76.17                                                  \\
PIP-DualP & 88.28                                                   & 87.06                                                & 83.78                                                  & 86.79    & 85                                                   & 82.13         \\ \hline                                                
\end{tabular}
\caption{Performance of the consolidated methods in CIFAR100 and TinyImageNet with smaller selected local clients (L), $^o$ and $^*$ denote the results obtained from the first 4 and 5 tasks due to crash.}
\label{tab:smaller_clients}
\end{table}
% \begin{figure}[h]
% \begin{center}
% %\framebox[4.0in]{$\;$}
% \includegraphics[width=0.5\textwidth]{n_clients_v2.png}
% \end{center}
% \caption{Performance of the consolidated methods in CIFAR100 and TinyImageNet with smaller local clients}
% \label{fig:smaller_clients}
% \end{figure}

% \vspace{-7pt}

\subsection{Complexity and Running Time Analysis}
% We evaluate the complexity of the proposed method as well as the complexity of the baseline method. 
% \vspace{-3pt}
Following the pseudo-code in Algorithm \ref{alg:pip},  PIP generates prototypes when its prototype set is empty after a local epoch of a round-$r$ (line 21), augment the prototypes in each local epoch (23), and updates the prototypes after local epochs (line 26-27). Knowing that generating prototypes from $\mathcal{T}^{r,t}_l$ costs $O(N_l^t)$, augmenting the prototypes costs $O(1)$ since it runs $m \in \{1..5\}$ times, then the PIP complexity will be: 

\begin{equation} \label{}
\small
    O(PIP) = O(1) + T.R_T. (O(\text{clients}) + O(\text{server}))
\end{equation}
\begin{equation} \label{}
\small
    O(PIP) = O(1) + T.R_T. (L.O(\text{1-client}) + O(\text{server}))
\end{equation}
\begin{equation} \label{}
\small
    O(PIP) = O(1) + T.R_T. (L.O(\text{1-client}) + O(1))
\end{equation}
\begin{equation} \label{}
\small
\begin{split}
    O(PIP) &= O(1) + T.R_T.L.O(E.(\sum_{b=1}^{\beta} N_{bl}^t+N_l^t)+O(N_l^t) \\
            &+ O(1)) + O(T.R_T)
\end{split}
\end{equation}
Since we have $\sum_{b=1}^{\beta} N_{bl}^t=N_l^t$, then we have
\begin{equation} \label{}
\small
\begin{split}
    O(PIP) &= O(1) + T.R_T.L.O(E.(N_l^t+N_l^t)+O(N_l^t) \\
           &+ O(1)) + O(T.R_T)
\end{split}
\end{equation}
\begin{equation} \label{}
\small
\begin{split}
    O(PIP) &= O(1) + T.R_T.L.O(E.(N_l^t)+O(N_l^t) \\
           &+ O(1)) + O(T.R_T)
\end{split}
\end{equation}
\begin{equation} \label{}
\begin{split}
     O(PIP) &=  O(1) + O(T.R_T.L.E.N_l^t)+O(T.R_T.L.N_l^t) \\
     &+ O(T.R_T.L) + O(T.R_T)
\end{split}
\end{equation}
\begin{equation} \label{}
\small
\begin{split}
    O(PIP) &= O(T.R_T.L.E.N_l^t)+O(T.R_T.L.N_l^t) \\
    & + O(T.R_T.L) + O(T.R_T)
\end{split}
\end{equation}
Substituting the equalities in the previous definition  that $R_T=R/T$ and $N_l=T.N_l^t=T$ the complexity of PIP will be
\begin{equation} \label{}
\small
\begin{split}
     O(PIP) &= O(T.R/T.L.E.N_l^t)+O(T.R/T.L.N_l^t) \\
     & + O(T.R/T.L) + O(T.R/T)
\end{split}
\end{equation}
\begin{equation} \label{}
\small
     O(PIP) = O(R.L.E.N_l^t)+O(R.L.N_l^t)+O(R.L) + O(R)
\end{equation}
\begin{equation} \label{}
\small
    O(PIP) = O(R.L.E.N_{l}^t)
\end{equation}
Since $N_{l}^t < N_{l} $ and E is set as a small constant in our method i.e. 2, then the PIP complexity will be:
\begin{equation} \label{}
\small
    O(PIP) = O(R.L.N_l)
\end{equation}

The baseline's complexity is presented in the supplemental document. Our complexity analysis shows that both the baseline method and our proposed method have the same complexity i.e. $O(R.L.N_l)$ where $R$ is the number of global rounds,  $L$ is the number of participating local clients in each round, and $N_l$ is the size of the training data in each client. 
% The detailed complexity analysis is provided in \textbf{Appendix H}.\ref{app:complexity}.
Table \ref{tab:training_time} summarizes the training time of the consolidated method in three datasets with T=10. The table shows that the proposed method requires lower training time and far lower communication cost than the current competing SOTAs i.e. LGA and GLFC in all datasets. The training time is higher than the baseline training time because in our proposed method there are additional processes that do not exist in the baseline i.e. prototype generation, augmentation, injection (concatenation with prompt features), aggregation, and feedback. TARGET may achieve the smallest running time, but It has poor performance with more than $60\%$ margin compared to our proposed method in all datasets.

\begin{table}[h]
% \vspace{-7pt}
% \setlength{\abovecaptionskip}{1pt plus 1pt minus 1pt}
% \setlength{\belowcaptionskip}{-15pt plus 0pt minus 0pt}
\centering
% \scriptsize
\small
\begin{tabular}{lcccc}
\hline
Method         & Com.Cost   & CIFAR100 & MiniImageNet & TinyImagenet \\ \hline
GLFC           & 61.7 MB    & 24.3h     & 36.6h         & 46.4h         \\ 
LGA            & 61.7 MB    & 23.2h     & 35.7h         & 45.9h        \\
TARGET            & 69.8 MB & 2.4h     & 2.3h         & 3.4h        \\ \hline
Fed-L2P         & 0.49 MB   & 5.1h        & 6.6h         & 15.3h         \\ 
Fed-DualP       & 1.32 MB & 11.3h      & 11.1h         & 19.5h         \\ 
PIP-L2P         & 0.52 MB & 7.5h      & 8.0h         & 17.8h        \\ 
PIP-DualP      & 1.35 MB & 13.0h     & 14.5h         & 25.5h        \\ \hline
\end{tabular}
\caption{Training time of the consolidated algorithm in CIFARR100, MiniImagenet and TinyImageNet.}
\label{tab:training_time}
\end{table}
% \vspace{-10pt}

\section{Concluding Remarks}
% \vspace{-3pt}
In this paper, we propose a new approach named prompt-based federated learning, new baselines named Fed-L2P and Fed-DualP, and a novel method named prototype-injected prompt (PIP) for the FCIL problem. PIP consists of three main ideas: a) prototype injection on prompt, b) prototype augmentation, and c) weighted Gaussian aggregation on the server side. Our experimental result shows that the proposed method outperforms the current SOTAs with a significant gap (up to $33\%$) in CIFAR100, MiniImageNet, and TinyImageNet datasets. Our extensive analysis demonstrates the robustness of our proposed method in different task sizes, smaller participating local clients, and smaller global rounds. Our proposed method has the same complexity as the baseline method and experimentally requires shorter training time than the current SOTAs. In practice, our proposed method can be applied in both cross-silo and cross-domain federated class incremental learning. Our future work is directed to federated few-shot class incremental learning where each client only holds a few samples for each task.

%%%%%%%%%%%%%%%%%%%%%%%%%%%%%%%%%%%%%%%%%%%%%%%%%%%%%%%%%
\section{Acknowledgement}
Muhammad Anwar Ma'sum acknowledges the support of Tokopedia-UI Centre of Excellence for GPU access to run the experiments. Savitha Ramasamy acknowledges the support of the National Research Foundation, Singapore under its AI Singapore Programme (AISG Award No: AISG2-RP-2021-027)

% \bibliography{samples/sample-base}
\bibliographystyle{ACM-Reference-Format}
\bibliography{main}

\newpage
\noindent \LARGE {\textbf{Supplemental document for  PIP: Prototypes-Injected Prompt for Federated Class Incremental Learning}}

\normalsize
This document offers supplementary materials for PIP: Prototypes-Injected Prompt for Federated Class Incremental Learning.  Section A explains the derivation of the weighted aggregation of PIP. Section B presents the algorithm of the proposed baselines. Section C elaborates on a theoretical analysis of the proposed method that proves the convergence and generalization of PIP. Section D presents a detailed analysis of baseline complexity. The rest presents the detailed numerical results in various scenarios. 

\received{20 February 2007}
\received[revised]{12 March 2009}
\received[accepted]{5 June 2009}

% \maketitle

% \end{frontmatter}

\appendix

\section{Derivation of Weighted Aggregation}\label{app:wagg}
Suppose that we have $n$ samples of an observation $x_i$ with the weight of $w_i$. The we mean and variance as:
\def\theequation{A\arabic{equation}}
\def\thetable{A\arabic{table}}
\begin{equation}\label{}
\small
    \mu =  \frac{1}{\sum_{i=1}^{n}w_i} \sum_{i=1}^{n}(x_i . w_i)
\end{equation} 
\begin{equation} \label{}
\small
 \begin{split}
    {\sigma}^2 &=  \frac{1}{\sum_{i=1}^{n}w_i} \sum_{i=1}^{n}w_i{(x_i - \mu)}^2 \\
    &\approx \frac{1}{\sum_{i=1}^{n}w_i} \sum_{i=1}^{n}(w_i{x_i}^2 - w_i{\mu}^2)
 \end{split}
\end{equation}
Or we have
\begin{equation} \label{}
\small
    {\sum_{i=1}^{n}w_i} {\sigma}^2=\sum_{i=1}^{n}(w_i{x_i}^2 - w_i{\mu}^2) =\sum_{i=1}^{n}w_i{x_i}^2 - \sum_{i=1}^{n}w_i{\mu}^2
\end{equation}
that equal
\begin{equation} \label{}
\small
    {\sum_{i=1}^{n}w_i} {\sigma}^2 + \sum_{i=1}^{n}w_i{\mu}^2= \sum_{i=1}^{n}w_i{x_i}^2
\end{equation}
If we have another $m$ observation, then we have
\begin{equation} \label{}
\small
    {\sum_{i=1}^{n+m}w_i} {\sigma}^2 + \sum_{i=1}^{n+m}w_i{\mu}^2 = \sum_{i=1}^{n+m}w_i{x_i}^2
\end{equation}
\begin{equation} \label{}
\small
    {\sum_{i=1}^{n+m}w_i} {\sigma}^2 + \sum_{i=1}^{n+m}w_i{\mu}^2 = \sum_{i=1}^{n}w_i{x_i}^2 + \sum_{i=n+1}^{n+m}w_i{x_i}^2
\end{equation}
\begin{equation} \label{}
\small
 \begin{split}
    {\sum_{i=1}^{n+m}w_i} {\sigma}^2 + \sum_{i=1}^{n+m}w_i{\mu}^2 =({\sum_{i=1}^{n}w_i} {\sigma_{1:n}}^2 + \sum_{i=1}^{n}w_i{\mu_{1:n}}^2) + \\ 
    ({\sum_{i=n+1}^{n+m}w_i} {\sigma_{n+1:m}}^2 + \sum_{i=n+1}^{n+m}w_i{\mu_{n+1:m}}^2)
\end{split}
\end{equation}
\begin{equation} \label{}
\small
 \begin{split}
    {\sum_{i=1}^{n+m}w_i} {\sigma}^2 + \sum_{i=1}^{n+m}w_i{\mu}^2 = {\sum_{i=1}^{n}w_i}  ({\sigma_{1:n}^2} + {{\mu_{1:n}}^2}) + \\
    {\sum_{i=n+1}^{n+m}w_i}  ({\sigma_{n+1:m}^2} + {{\mu_{n+1:m}}^2})
 \end{split}
\end{equation}

Therefore:
\begin{equation} \label{}
\small
\begin{split}
    {\sigma}^2 =& \frac{{\sum_{i=1}^{n}w_i}  ({\sigma_{1:n}^2} + {{\mu_{1:n}}^2})  } {{\sum_{i=1}^{n+m}w_i}} + \\
    & \frac{{\sum_{i=n+1}^{n+m}w_i}  ({\sigma_{n+1:m}^2} + {{\mu_{n+1:m}}^2}) - \sum_{i=1}^{n+m}w_i{\mu}^2}{{\sum_{i=1}^{n+m}w_i}} 
        % {\sigma}^2 = \frac{{\sum_{i=1}^{n}w_i}  ({\sigma_{1:n}^2} + {{\mu_{1:n}}^2}) + {\sum_{i=n+1}^{n+m}w_i}  ({\sigma_{n+1:m}^2} + {{\mu_{n+1:m}}^2}) - \sum_{i=1}^{n+m}w_i{\mu}^2 }{{\sum_{i=1}^{n+m}w_i}}
\end{split}
\end{equation}
\begin{equation} \label{}
\small
\begin{split}
    {\sigma}^2 =& \frac{ {\sum_{i=1}^{n}w_i}  ({\sigma_{1:n}^2} + {{\mu_{1:n}}^2})} {{\sum_{i=1}^{n+m}w_i}} + \\
    & \frac{{\sum_{i=n+1}^{n+m}w_i}  ({\sigma_{n+1:m}^2} + {{\mu_{n+1:m}}^2})} {{\sum_{i=1}^{n+m}w_i}} - {\mu}^2 
\end{split}
    % {\sigma}^2 = \frac{ {\sum_{i=1}^{n}w_i}  ({\sigma_{1:n}^2} + {{\mu_{1:n}}^2}) + {\sum_{i=n+1}^{n+m}w_i}  ({\sigma_{n+1:m}^2} + {{\mu_{n+1:m}}^2})} {{\sum_{i=1}^{n+m}w_i}} - {\mu}^2 
\end{equation}
% $w_{lc}^t \sim \mathcal{N}(\mu_{lc}^t,\,{\sigma_{lc}^t}^{2})$

The derivation above shows that if we have two weighted Gaussian distributions e.g. $X_1 \sim \mathcal{N}(\mu_1,\sigma_1^{2})$ and $X_2 \sim \mathcal{N}(\mu_2,\sigma_2^2)$  with total weight $W_1=\sum_{i=1}^{|X_1|}w_i$ and $W_2=\sum_{j=1}^{|X_2|}w_j$ respectively, then the aggregated distribution will be:
\begin{equation}
\small
    \mu* =  \frac{(\mu_1.W_1 + \mu_2.W_2)}{W_1+W_2}
\end{equation} 
\begin{equation}
\small
    {\sigma*}^2 =  \frac{(({\mu_1}^2+{\sigma_1}^2).W_1 + ({\mu_2}^2+{\sigma_2}^2).W_2)}{W_1+W_2} - {\mu*}^2
\end{equation}
Generalizing equations above into $N$ observations i.e. $X_1 \sim \mathcal{N}(\mu_1,{\sigma_1}^2)$, $X_2 \sim \mathcal{N}(\mu_2,{\sigma_2}^2)$, ... $X_N \sim \mathcal{N}(\mu_N,{\sigma_N}^2)$  with total weight $W_1$, $W_2$, ...  $W_N$ respectively. then the aggregated distribution will be:
\begin{equation}
\small
    \mu* =  \frac{\sum_{i=1}^{N}(\mu_i.W_i)}{\sum_{i=1}^{N}W_i}
\end{equation} 
\begin{equation}
\small
    {\sigma*}^2 =  \frac{\sum_{i=1}^{N}({\mu_i}^2+{\sigma_i}^2).W_i} {\sum_{i=1}^{N}W_i} - {\mu*}^2
\end{equation}

\section{Baseline Algorithm}\label{app:pip}
We present the detailed algorithm of our 
% proposed method in Algorithm \ref{alg:pip}, and detailed algorithm of the 
proposed baseline in Algorithm \ref{alg:baseline}.

\begin{algorithm*}
% \small
\caption{Fed-Prompt}\label{alg:baseline}
\begin{algorithmic}[1]
% \Require $n \geq 0$
\State \textbf{Input:} Number of clients $N$, number of selected local clients $L$, total number of rounds $R$, number of task $T$, local epochs $E$,  batch size $B$.
% \KwInput{Number of clients $N$, number of selected local clients $L$, total number of rounds $R$, number of task $T$, local epochs $E$,  batch size $B$, number}
% \item[\hspace{3 mm} // Some label if needed]
\State Distribute frozen ViT backbone $f$ to all clients $\{S_l\}_{l=1}^{N}$ and central server $S_G$ 
\State Initiate prompt, key, and head layer for all clients and central server $p_G = p_l$, $\phi_G = \phi_l,\, l \in \{1..N\}$ 
\State $R_T \leftarrow R/T$, $R_T$ represents round per task
\For{$t=1:T$} 
    \For{$r=1:R_T$}
        \State $S_l \leftarrow$ randomly select $L$ local clients from $N$ total clients
        \State \textbf{Clients execute}: 
        \State Receive global parameters i.e. prompt and head layer $p_G$ and $\phi_G$ 
        \State Assign local parameters $(p_l,\phi_l) \leftarrow$  $(p_G,\phi_G)$ 
        \State $\mathcal{B} \leftarrow$ Split $\mathcal{T}_l^t$ into $B$ sized batches
        \For{$e=1:E$}
            \For{$b=1:\mathcal{B}$}
                \State Compute prompt-generated feature $f_{p_l}(x)$ as in Eq. (1) to (4)
                \State Compute logits $f_{\phi_l}(f_{p_l}(x))$
                \State Compute loss $\mathcal{L}_{t}$ as in Eq. (5)
                \State Update local parameters $(p_l,\phi_l)$ based on $\mathcal{L}_{t}$ as in Eq. (6)  
            \EndFor
        \EndFor
        \State Store local parameters $(p_l,\phi_l)$
        \State Send local parameters $(p_l,\phi_l)$ to server
        \State \textbf{Server executes}:
        \State Receives selected local clients $S_l$ parameters $(p_l,\phi_l)$ for $l \in [1..L]$
        \State Aggregates clients' parameters into global parameters  $(p_G,\phi_G) \leftarrow \frac{1}{L}\sum_{l=1}^L(p_l,\phi_l)$   
        \State Send global parameters  $(p_G,\phi_G)$ to clients for the next round  
    \EndFor
\EndFor
\State \textbf{Output:}  Global parameters $(p_G,\phi_G))$ and local parameters $(p_l,\phi_l), l \in {1..N}$  
\end{algorithmic}
\end{algorithm*}

\section{Detailed Theoretical Analysis}\label{app:baseline}
Let $\theta=(p,\phi)$ is the trainable parameters and $F(\theta)=\mathbb{E}[\mathcal{L}_{l+}(\mathcal{T};\theta)]=\mathbb{E}[\mathcal{L}_{l+}(\mathcal{T};(p,\phi))]$ is the expected loss function, $k, E$, $R$, and $L_S$ is local iteration, local epoch, global round, and number of selected local clients respectively. Please note that in this analysis, $L_S$ denotes the number of selected local clients, while $L \geq 1$ denotes a constant for $L$-smooth coefficient. We adopt the following assumptions adapted from SGD optimization convergence analysis \cite{bottou2018optimization} and FedAvg convergence analysis \cite{li2019convergence}.

\noindent\textbf{Assumption 1:} $F_1,...F_l,...,F_{L_S}$ are all $L-$smooth: for all $\theta$ and $\theta'$, $F_l(\theta) \leq F_l(\theta') + (\theta - \theta')^T \nabla F_l(\theta) + \frac{L}{2} ||\theta-\theta'||_2^2.$

\noindent\textbf{Assumption 2:} $F_1,...F_l,...,F_{L_S}$  are all $\mu-$strongly convex: for all $\theta$ and $\theta'$, $F_l(\theta) \leq F_l(\theta') + (\theta - \theta')^T \nabla F_l(\theta) + \frac{\mu}{2} ||\theta-\theta'||_2^2.$

\noindent\textbf{Assumption 3:} Let $\xi_l^k$ be the random uniformly sampled from $l$-th local data at $k-th$ iteration . The variance of stochastic gradients in each client is bounded by: $\mathbb{E}||\nabla F_l(\theta_l^k, \xi_l^k) - \nabla F_l(\theta_l^k)|| \leq \sigma_l^2$ for $l=1,2,..., L_S$

\noindent\textbf{Assumption 4:}The expected squared norm of stochastic gradients in each client is bounded by: $\mathbb{E}||\nabla F_l(\theta_l^k, \xi_l^k)|| \leq G^2$ for all $l=1,2,...,L_S$ and $k=1,2,...., K$ where $K \in \mathbb{N}$.

\noindent\textbf{Assumption 5:} $\sum_{k=1}^{\infty} \alpha_l^k = \infty$ and $\sum_{k=1}^{\infty} {\alpha_l^k}^2 < \infty$ where  $\alpha_l^k$ is the learning rate of $l-th$ client in $k$-th step training.

\subsection{Proof of Theorem 1}
Suppose that a client $S_l$ is trained locally with its local data $\mathcal{T}_l^t \cup Z^t$, where $\mathcal{T}_l^t$ is local training samples for $t$-th task and $Z^t$ is shared prototype for $t$-th task respectively. We already stated that $Z^t$ contains augmented prototype so that $|z_{c_2}^t| \approx |x_{c_1}|$ for $z_{c_2}^t \in Z^t$ and $x_{c_1}^t \in \mathcal{T}_{l_c}^t, {T}_{l_c}^t$ is subset of $\mathcal{T}_l^t$ where $y_i=c$ so the number of prototype of unavailable classes in $\mathcal{T}_l^k$ and the samples of available classes in $\mathcal{T}_l^k$ are balanced. Then the local model $\theta_l = (p_l, \phi_l)$ is updated in $K$ iteration based on sample $\xi_l^k$ that sampled (e.g. minibatch) from $\mathcal{T}_l^t \cup Z^t$. Since the backbone (feature extractor) is frozen, and $\mathcal{T}_l^t \cup Z^t$ has balance samples for all classes, then $\xi_l^k$ approximates  $\xi^k$  that is a sample from  $\mathcal{T}^t$. The local model is updated by the stochastic gradient (SG) approach as presented in equations (6) and (10) in the main paper. Suppose that $g(\theta_l, \xi_l^k)$ is the stochastic gradient function, then the update process can be simplified as:
\begin{equation} \label{}
\small
    \theta_l^{k+1} \leftarrow \theta_l^{k} - \alpha_l^k g(\theta_l^k, \xi_l^k)
\end{equation}
Under assumption 1, and local training updates $\theta$ by iterating SG with sample $\xi_l^k$, then we have:
\begin{equation} \label{}
\small
\begin{split}
     F_l(\theta_l^{k+1}) - F_l(\theta_l^{k}) &\leq (\theta_l^{k+1} - \theta_l^{k})^T \nabla F_l(\theta_l^k) + \frac{L}{2} ||\theta_l^{k+1} - \theta_l^{k}||_2^2 \\
     &\leq \nabla F_l(\theta_l^k)^T(\theta_l^{k+1} - \theta_l^{k}) + \frac{L}{2} ||\theta_l^{k+1} - \theta_l^{k}||_2^2 \\
     &\leq -\alpha_l^k \nabla F_l(\theta_l^k)^T g(\theta_l^k,\xi_l^k) + {\alpha_l^k}^2\frac{L}{2} ||g(\theta_l^k,\xi_l^k)||_2^2
\end{split}
\end{equation}
The equation above can be derived into: 
\begin{equation} \label{}
\small
\begin{split}
     \mathbb{E}_{\xi_l^k}[F_l(\theta_l^{k+1})] - F_l(\theta_l^{k}) \leq & -\alpha_l^k  \nabla F_l(\theta_l^k)^T \mathbb{E}[g(\theta_l^k,\xi_l^k)] \\
     & + {\alpha_l^k}^2\frac{L}{2} \mathbb{E}_{\xi_l^k}[||g(\theta_l^k,\xi_l^k)||_2^2] 
\end{split}
\end{equation}

The inequation above shows how SG optimizes $\theta_l^k$ at a step $k$, and we can see clearly that the decreasing of $F_l$ (left side) is bounded by a quantity in the right side involving $\nabla F_l$ which is directional derivative of $F_l$ at $\theta_l^k$ along with  $-g(\theta_l^k, \xi_l^k)$ (first term)  and second moment of $g(\theta_l^k, \xi_l^k)$ (second term). In the case where $g(\theta_l^k, \xi_l^k)$ is unbiased estimator of $\nabla F_l$, then the inequation above can be derived as:

\begin{equation} \label{}
\small
     \mathbb{E}_{\xi_l^k}[F_l(\theta_l^{k+1})] - F_l(\theta_l^{k}) \leq -\alpha_l^k \nabla ||F_l(\theta_l^k)||_2^2 + {\alpha_l^k}^2\frac{L}{2} \mathbb{E}_{\xi_l^k}[||g(\theta_l^k,\xi_l^k)||_2^2]
\end{equation}

The inequation above shows that as long as the stochastic directions and stepsize are chosen, then the SGD convergence is guaranteed. To avoid the harm of the second term of the right side in the inequation above, we apply the restriction below 
\begin{equation} \label{}
\small
      \mathbb{V}[g(\theta_l^k,\xi_l^k)] = \mathbb{E}[||g(\theta_l^k,\xi_l^k)||_2^2] - || \mathbb{E}[g(\theta_l^k,\xi_l^k)] ||_2^2.
\end{equation}

Adopting first and second-moment limit as in \cite{bottou2018optimization}, then we add the following assumption.

\noindent\textbf{Assumption 6:} The objective function $F_l$ and SG satisfy the following conditions.

\noindent(a). The sequence of $\{\theta_l^k\}$ is contained in an open space where $F_l$ is bounded below by a scalar $F_{inf}$

\noindent(b) Exist scalars $\nu_G \geq \nu > 0$ so that for all $k \in \mathbb{N}$ satisfy:
\begin{equation} \label{}
\small
\begin{split}
     &\nabla F_l(\theta_l^k)^T \mathbb{E}_{\xi_l^k}[g(\theta_l^k,\xi_l^k)] \geq \nu||\nabla F_l(\theta_l^k)T||_2^2, and \\
     &|| \mathbb{E}_{\xi_l^k}[g(\theta_l^k,\xi_l^k)]||_2 \leq \nu_G||\nabla F_l(\theta_l^k)||_2.
\end{split}
\end{equation}
% If $g(\theta_l^k,\xi_l^k)$, is unbiased estimator of $\nabla F_l(\theta_l^k)$ then It satisfies $\mu_G=\mu=0$.

\noindent(c) Exist scalars $m_1 \geq 0$ and $m_2 \geq 0$ so that for all $k \in \mathbb{N}$ satisfy:
\begin{equation} \label{}
\small
 \mathbb{V}[g(\theta_l^k,\xi_l^k)] \leq m_1 + m_2 ||\nabla F_l(\theta_l^k)||_2^2
\end{equation}

\noindent Combining assumption 6 and restriction criteria as presented in equation (A19), then we have:
\begin{equation} \label{}
\begin{split}
\mathbb{E}_{\xi_l^k}[||g(\theta_l^k,\xi_l^k)||_2^2] \leq m_1 + m_G ||\nabla F_l(\theta_l^k)||_2^2, with \\
m_G = m_2+\nu_G^2 \geq \nu^2 > 0
\end{split}
\end{equation}

\noindent Then by substituting $\mathbb{E}_{\xi_l^k}[||g(\theta_l^k,\xi_l^k)||_2^2]$ from equation (A22) into equation (A17), we have:

\begin{equation} \label{}
\small 
\begin{split}
         \mathbb{E}_{\xi_l^k}[F_l(\theta_l^{k+1})] - F_l(\theta_l^{k}) \leq & -\alpha_l^k\nabla F_l(\theta_l^k)^T \mathbb{E}[g(\theta_l^k,\xi_l^k)] \\ 
         & + {\alpha_l^k}^2\frac{L}{2} (m_1 + m_G ||\nabla F_l(\theta_l^k)||_2^2)
\end{split}
\end{equation}
\noindent Assumption 5 ensures that $\{\alpha_l^k\}\rightarrow0$ practically can be achieved by using learning rate decay that reduces learning rate in every step of local training. Then by choosing $\alpha_l^kLm_G \leq \nu$ and substituting $\nabla F_l(\theta_l^k)^T \mathbb{E}[g(\theta_l^k,\xi_l^k)] $ in equation (A23) with the condition in assumption 6.b, we have
\begin{equation} \label{}
\small 
\begin{split}
         \mathbb{E}_{\xi_l^k}[F_l(\theta_l^{k+1})] - F_l(\theta_l^{k}) \leq & -\alpha_l^k\nu||\nabla F_l(\theta_l^k)||_2^2 \\ 
         & + {\alpha_l^k}^2\frac{L}{2} (m_1 + m_G ||\nabla F_l(\theta_l^k)||_2^2) 
\end{split}
\end{equation}
Applying expectation to the equation above yield
 \begin{equation} \label{}
\small 
\begin{split}
         \mathbb{E}_{\xi_l^k}[F_l(\theta_l^{k+1})] - \mathbb{E}[F_l(\theta_l^{k})] \leq & -\alpha_l^k\nu\mathbb{E}[||\nabla F_l(\theta_l^k)||_2^2] \\ 
         & + {\alpha_l^k}^2\frac{1}{2} (m_1 + m_G \mathbb{E}[||\nabla F_l(\theta_l^k)||_2^2]) \\
         \mathbb{E}_{\xi_l^k}[F_l(\theta_l^{k+1})] - \mathbb{E}[F_l(\theta_l^{k})] \leq & - (\nu-\frac{1}{2}\alpha_l^kLm_G) \alpha_l^k\mathbb{E}[||\nabla F_l(\theta_l^k)||_2^2] \\ 
         & + \frac{1}{2} {\alpha_l^k}^2Lm_1 \\
         \mathbb{E}_{\xi_l^k}[F_l(\theta_l^{k+1})] - \mathbb{E}[F_l(\theta_l^{k})] \leq & -\frac{1}{2}\nu\alpha_l^k\mathbb{E}[||\nabla F_l(\theta_l^k)||_2^2] \\
         & + \frac{1}{2} {\alpha_l^k}^2Lm_1
\end{split}
\end{equation}
\noindent Sum both sides for $k \in \{1,...,K\}$ we get
 \begin{equation} \label{}
\small 
\begin{split}
F_{inf} - \mathbb{E}[F(\theta_l^1)] &\leq \mathbb{E}[F_l(\theta_l^{K+1})] - \mathbb{E}[F_l(\theta_l^{1})] \\ 
F_{inf} - \mathbb{E}[F(\theta_l^1)]  &\leq -\frac{1}{2}\nu\sum_{k=1}^{K}\alpha_l^k\mathbb{E}[||\nabla F_l(\theta_l^k)||_2^2] + \frac{1}{2} Lm_1 \sum_{k=1}^{K}{\alpha_l^k}^2
\end{split}
\end{equation}

Dividing by $\nu$ for both sides, then we get
\begin{equation} \label{}
\small 
\sum_{k=1}^{K}\alpha_l^k\mathbb{E}[||\nabla F_l(\theta_l^k)||_2^2] \leq \frac{2(\mathbb{E}[F(\theta_l^1)]-F_{inf})}{\nu} + \frac{Lm_1}{\nu} \sum_{k=1}^{K}{\alpha_l^k}^2
\end{equation}
Applying $\lim_{K\rightarrow \infty}$ and assumption 5 to the equation above we get
\begin{equation} \label{}
\small 
\begin{split}
\lim_{K\rightarrow \infty}\sum_{k=1}^{K}\alpha_l^k\mathbb{E}[||\nabla F_l(\theta_l^k)||_2^2] \leq& \frac{2(\mathbb{E}[F(\theta_l^1)]-F_{inf})}{\nu} \\
& +\frac{Lm_1}{\nu} \lim_{K\rightarrow \infty}\sum_{k=1}^{K}{\alpha_l^k}^2 < \infty
\end{split}
\end{equation}
Dividing both sides with $\sum_{k=1}^{K}{\alpha_l^k}$, and following assumption 5 where $\lim_{K\rightarrow \infty}\sum_{k=1}^{K}{\alpha_l^k} = \infty$ and $\lim_{K\rightarrow \infty}\sum_{k=1}^{K}{\alpha_l^k}^2 < \infty$, then the right side will return 0. Therefore, we have
\begin{equation} \label{}
\small 
\lim_{K\rightarrow \infty} \frac{\sum_{k=1}^{K}\mathbb{E}[\alpha_l^k ||\nabla F_l(\theta_l^k)||_2^2]}{\sum_{k=1}^{K}{\alpha_l^k} } = 0
\end{equation}
\begin{equation} \label{}
\small 
\lim_{K\rightarrow \infty} \mathbb{E}[\frac{\sum_{k=1}^{K}\alpha_l^k||\nabla F_l(\theta_l^k)||_2^2}{\sum_{k=1}^{K}{\alpha_l^k} }] = 0
\end{equation}
\begin{equation} \label{}
\small 
\lim_{k\rightarrow \infty} \mathbb{E}[||\nabla F_l(\theta_l^k)||_2^2] = 0
\end{equation}
The equation above presents the convergence for local training in $l$-th client where the gradient of loss $F$ converges to 0 along with the increase of training step/iteration $k$. 

\subsection{Proof of Theorem 2}
Suppose that set of local client $\{S_l\}_{l=1}^{l=L_S}$ is trained locally with its local data $\{\mathcal{T}_l^t \cup Z^t\}_{l=1}^{l=L_S}$, where $\mathcal{T}_l^t$ is local training sample for $l$-th client. Local training is conducted in $k$ steps/iterations using a sample i.e. minibatch of local training set $\xi_l^k \in \mathcal{T}_l^t$ on each step. Global synchronization is executed in each round $r = \{1,2,...,R\}$. We global synchronization step as $\mathcal{I}_E = \{rE|r=1,2,...R\}$. Following \cite{li2019convergence} we define $\theta_l^{k+1}$ represents the local parameter of $l$-client after communication steps, while $\vartheta_l^{k+1}$ represents the local parameter after an immediate result of one step SGD. Therefore the definition satisfies:

\begin{equation} \label{}
\small
    \vartheta_l^{k+1} = \theta_l^{k} - \alpha_l^k \nabla F_l(\theta_l^k, \xi_l^k)
\end{equation}
\begin{equation} \label{}
\small
\theta_l^{k+1} =
\begin{cases}
    \vartheta_l^{k+1} \; \text{if } k+1\notin \mathcal{I}_E\\
    \sum_{l=1}^{L_S} w_l^k \vartheta_l^{k+1}  \; \text{if } k+1\in \mathcal{I}_E\\
\end{cases}
\end{equation}

\noindent Where $w_l = \omega_l/\sum_{l=1}^{L_S} \omega_l, \text{where } \omega_l$ is the weight of $l$-th client. We define $\bar{\vartheta}_l^{k+1} = \sum_{l=1}^{L_S} w_l {\vartheta}_l^{k+1}$ and $\bar{\theta}_l^{k+1} = \sum_{l=1}^{L_S} w_l {\theta}_l^{k+1}$, $\bar{\vartheta}_l^{k+1}$ is the result of single step SGD iteration from $\bar{\theta}_l^{k+1}$. We also define $\bar{g}^k = \sum_{l=1}^{L_S} w_l \nabla F_l(\theta_l^k)$ and ${g}^k = \sum_{l=1}^{L_S} w_l \nabla F_l(\theta_l^k, \xi_l^k)$. 
We adopt the following lemmas from \cite{li2019convergence} where derived from fully participating clients in federated learning. 

\noindent \textbf{Lemma 1:} By applying assumptions 1 and 2, in one step SGD training and chose $\alpha \leq \frac{1}{4L}$ we have

\noindent $\mathbb{E}[||\bar{\vartheta}^{k+1} - \theta^*||^2] \leq (1 - \alpha^k\mu)\mathbb{E}[||\bar{\theta}^{k} - \theta^*||^2]  - (\alpha^k)^2\mathbb{E}[||{g}^{k} - \bar{g}^k||^2] + 6L(\alpha^k)^2\Gamma +2\mathbb{E}[\sum_{l=1}^{L_S}w_l||\bar{\theta}^{k} - \theta_l^k||^2]$ where $\Gamma = F^* - \sum_{l=1}^{L_S}w_lF_l^* \geq 0$.

\noindent \textbf{Lemma 2:} By applying assumption 3, the gradient function follows:  

\noindent $\mathbb{E}[||\bar{g}^{k} - \bar{g}^k||^2] \leq \sum_{l=1}^{L_S}w_l^2 \sigma_l^2$, where $\sigma_l^2$ is the variance of $\theta_l$

\noindent \textbf{Lemma 3:} By applying assumption 4, where $\alpha^k$ is non-increasing and it satisfies $\alpha^k \leq \alpha^{k+E} \text{for all} k \geq 0$, then we have
$\mathbb{E}[\sum_{l=1}^{L_S}||\bar{\theta}^{k+1} - \theta_l^k||^2]  \leq 4(\alpha^k)^2(E-1)^2G^2$

In fully participating clients we always have $\bar{\theta}^{k+1} = \bar{\vartheta}^{k+1}$. However, in partially participating clients we use a random sampling mechanism so that It satisfies 
$\mathbb{E}_{S_L}[\bar{\theta}^{k+1}] = \bar{\vartheta}^{k+1}$. We also adopt the bounding condition from \cite{li2019convergence} as shown in lemma 4. 

\noindent \textbf{Lemma 4:} The expected different between $\bar{\theta}^{k+1} \text{and } \bar{\vartheta}^{k+1}$ bounded by :  $\mathbb{E}_{S_L}[||\bar{\vartheta}^{k+1} - \bar{\theta}^{k+1}||^2]  \leq \frac{4}{L_S}(\alpha^k)^2E^2G^2$ and in the case of $w_l$ is uniform for all $l$-th client, then $\mathbb{E}_{S_L}[||\bar{\vartheta}^{k+1} - \bar{\theta}^{k+1}||^2]  \leq \frac{4(N_S-L_S)}{N_S-1}(\alpha^k)^2E^2G^2$, where $N_S$ is total clients and $L_S$ is number of selected clients.

\noindent Please note that 
\begin{equation} \label{}
\small
   ||\bar{\theta}^{k+1} - \theta^*||^2 = ||\bar{\theta}^{k+1}  - \theta^*||^2
\end{equation}
\begin{equation} \label{}
\small
   ||\bar{\theta}^{k+1} - \theta^*||^2 = ||\bar{\theta}^{k+1} - \bar{\vartheta}^{k+1} +  -\bar{\vartheta}^{k+1} - \theta^*||^2
\end{equation}
\begin{equation} \label{}
\small
\begin{split}
       ||\bar{\theta}^{k+1} - \theta^*||^2 =& ||\bar{\theta}^{k+1} - \bar{\vartheta}^{k+1}||^2 +  ||\bar{\vartheta}^{k+1} - \theta^*||^2 \\
       & + 2 ||\bar{\theta}^{k+1} - \bar{\vartheta}^{k+1}|| .||\bar{\vartheta}^{k+1} - \theta^*||
\end{split}
\end{equation}

\begin{equation} \label{}
\small
\begin{split}
       ||\bar{\theta}^{k+1} - \theta^*||^2 =& ||\bar{\theta}^{k+1} - \bar{\vartheta}^{k+1}||^2 +  ||\bar{\vartheta}^{k+1} - \theta^*||^2 \\
       & + 2 \langle \bar{\theta}^{k+1} - \bar{\vartheta}^{k+1} , \bar{\vartheta}^{k+1} - \theta^* \rangle
\end{split}
\end{equation}

In the case of $k+1 \notin \mathcal{I}_E$, then the term $||\bar{\theta}^{k+1} - \bar{\vartheta}^{k+1}||^2$ vanishes. Then by applying lemma 4, we get
\begin{equation} \label{}
\small
   \mathbb{E}[||\bar{\theta}^{k+1} - \theta^*||^2] \leq (1-\alpha^k\mu)\mathbb{E}[||\bar{\theta}^{k+1}  - \theta^*||^2] + (\alpha^k)B
\end{equation}
In the case of $k+1 \in \mathcal{I}_E$, then by applying lemma 4, we get

\begin{equation} \label{}
\small
   \mathbb{E}[||\bar{\theta}^{k+1} - \theta^*||^2] \leq (1-\alpha^k\mu)\mathbb{E}[||\bar{\theta}^{k+1}  - \theta^*||^2] + (\alpha^k) (B+C)
\end{equation}

where $B=\sum_{l=1}^{L_S}w_l\sigma_l^2 + 6L\Sigma+ 8(E-1)^2G^2$ and $C=\frac{4(N_S-L_S)}{N_S-1}(E^2G^2)$ if $w_l$ is uniform and $C=\frac{4}{L_S}(E^2G^2)$ otherwise.

By choosing $\alpha^k = \frac{\beta}{k+\delta}$ for some $\beta > 1/\mu$ and $\delta > 0$ so that $\alpha^1 \leq \min\{1/\mu,1/4L\}=1/4L$ and $\alpha^k \leq 2\alpha^{k+E}$ then we have $\mathbb{E}[||\bar{\theta}^{k+1} - \theta^*||^2] \leq \frac{v}{\delta+k}$ where $v=\max\{\frac{\beta^2(B+C)}{\beta\mu-1},(\delta+1)||\bar{\theta}^{k+1} - \theta^*||^2\}$

Then, by applying a strong convexity assumption of $F$ we have
\begin{equation} \label{}
\small
   \mathbb{E}[\bar{\theta}^{k}] - F^*  \leq \frac{L}{2}\Delta^k \leq \frac{L}{2}\frac{v}{\delta+k} 
\end{equation}
where $F^*$ is the minimum value of $F$ where optimum parameter $\theta^*$ is achieved. Later on, if we choose $\beta = 2/\mu, \delta=\max\{8L/\mu,E\} \text{and denote } \kappa = L/\mu, \alpha^k = 2/u (1/(\delta+k))$ then we have
\begin{equation} \label{}
\small
   \mathbb{E}[F(\bar{\theta}^k)]-F^* \leq \frac{\kappa}{(\delta+k-1)}(\frac{2(B+C)}{\mu} +\frac{\mu\delta}{2}\mathbb{E}||\theta^1-\theta^*||) 
\end{equation}
The equation generalizes federated learning where the model is trained in a total of $k$ steps/iterations where practical implementation satisfies $k = b.E.R, b$ is the number of batches. here we know that $k > R$ as $E \text{and} b$ are positive integers. Therefore, substituting $k$ with $R$ in the inequation above will produce a higher amount of the right side. Therefore, the inequation above can be generalized into:
\begin{equation} \label{}
\small
   \mathbb{E}[F(\theta^R)]-F^* \leq \frac{\kappa}{(\delta+R-1)}(\frac{2(B+C)}{\mu} +\frac{\mu\delta}{2}\mathbb{E}||\theta^1-\theta^*||) 
\end{equation}

The inequation above guarantees that to achieve a convergence condition in federated learning with weighted aggregation is upper bounded by the amount on the right side. The inequation above also shows the convergence guarantee by the method in the case of $b=1$ or $E=1$ or both, which simulates a small number of samples so that local training is only executed with 1 batch, or in an urgent (need to be fast) condition where local training is only executed with 1 epoch, or those two conditions occur in the same time.

\subsection{Proof of Theorem 3}
Given $\theta^*$ and $\theta$ are optimal parameter in $\mathcal{T}^t_l \cup Z^t$ and $\mathcal{T}^t_l$ respectively, where $\mathcal{T}^{t}_{l} \subset \mathcal{T}^{t}$, where  $|\mathcal{T}^{t}_{l}| / |\mathcal{T}^{t}| = \eta \in (0,1)$, then we have
\begin{equation} \label{}
\small
F(\theta;\mathcal{T}^t) =  \eta F(\theta;\mathcal{T}_l^t) + (1-\eta) F(\theta;(\mathcal{T}^t-\mathcal{T}_l^t))
\end{equation}
\begin{equation} \label{}
\small
F(\theta^*;\mathcal{T}^t) =  \eta F(\theta^*;\mathcal{T}_l^t) + (1-\eta) F(\theta^*;(\mathcal{T}^t-\mathcal{T}_l^t))
\end{equation}
Suppose that $\theta^o$ is the initial value both for $\theta \text{and } \theta^*$ that set by random uniform initiation method. Therefore for all class $c \in \mathcal{T}^t_{c} = \mathcal{T}^t_{y=c}$ It satisfy  $F(\theta^o;\mathcal{T}_c^t) = e^o$. After optimally learning on  $\mathcal{T}^t_l$ and $\mathcal{T}_l^t \cup Z^t$ then $\theta^o$ become to $\theta$ and $\theta^*$ respectively. Please note that $\theta$ learns only available class in $\mathcal{T}^t_l$, while $\theta^*$ learns classes that available in $\mathcal{T}^t_l$ and classes in $\mathcal{T}^t - \mathcal{T}^t_l$ via $Z^t$. Suppose that the loss for predicting classes in $\mathcal{T}^t_l$ is defined as $e^a < e^o$ then we have $F(\theta;\mathcal{T}_l^t) = F(\theta^*;\mathcal{T}_l^t) = e^a < e^o$. Since the backbone is frozen, $\theta^*$ learn $Z^t$ then we have $F(\theta;(\mathcal{T}^t - \mathcal{T}_l^t)) = e^o$, while $F(\theta^*;(\mathcal{T}^t - \mathcal{T}_l^t)) = e^b$, where $e^a \geq e^b \geq e^o$.

Then equation (A43) and (A44) can be derived to 
\begin{equation} \label{}
\small
F(\theta;\mathcal{T}^t) =  \eta e^a+ (1-\eta) e^o
\end{equation}
\begin{equation} \label{}
\small
F(\theta^*;\mathcal{T}^t) =  \eta e^a+ (1-\eta) e^b
\end{equation}
Substracting the equations above, then we have
\begin{equation} \label{}
\small
F(\theta;\mathcal{T}^t) - F(\theta^*;\mathcal{T}^t) =   \eta e^a+ (1-\eta) e^o - (\eta e^a+ (1-\eta) e^b)
\end{equation}
\begin{equation} \label{}
\small
F(\theta;\mathcal{T}^t) - F(\theta^*;\mathcal{T}^t) =   \eta e^a+ (1-\eta) e^o -\eta e^a -(1-\eta) e^b
\end{equation}
\begin{equation} \label{}
\small
F(\theta;\mathcal{T}^t) - F(\theta^*;\mathcal{T}^t) =  (1-\eta) e^o -(1-\eta) e^b
\end{equation}
\begin{equation} \label{}
\small
F(\theta;\mathcal{T}^t) - F(\theta^*;\mathcal{T}^t) =  (1-\eta) (e^o-e^b)
\end{equation}

Since $0 < \eta < 1 $ and $e^o > e^b$ the right side of the inequation above has a positive value. By choosing 
a small positive value $\epsilon > 0 $ where  $ (1-\eta) (e^o-e^b) \geq \epsilon $ then we have.

\begin{equation} \label{}
\small
F(\theta;\mathcal{T}^t) - F(\theta^*;\mathcal{T}^t) \geq  \epsilon
\end{equation}
Inequation above proves that $\theta^*$ is more generalized to $\mathcal{T}^t$ than $\theta$. This shows that our idea i.e. injecting the shared prototypes improves model generalization.

\section{Complexity Analysis}\label{app:complexity}
In this section, we analyze the complexity of our proposed baseline. Suppose that $N_l$ is the total samples of a dataset of a client across all tasks, $t \in\{1,...,T\}$ is task index, $N_l^t=|\mathcal{T}^{r,t}_{l}|=|\mathcal{T}^{t}_{l}|$  is the number of samples on task $t$in client-$l$ that satisfy $\mathcal{\sum}_{t=1}^{T} N_l^t = N_l$, $R$ is the total rounds of federated learning, $E$  is the number of local epoch for each client training. $\beta$ is the number of batches on each task that satisfy $\sum_{b=1}^{\beta} N_{bl}^t = N_l^t$. We simplify the derivation by analyzing the complexity in a common case that the tasks are divided evenly, therefore we have $|\mathcal{T}^1_l|=|\mathcal{T}^2_l|...=|\mathcal{T}^t_l|...=|\mathcal{T}^T_l|$, that equal $N_l=T.N_l^t=T.|\mathcal{T}_l^t|$. Let $O(.)$ denote the complexity of a process.

\textbf{Baseline Complexity:}Following the pseudo-code in Algorithm \ref{alg:baseline} then we have

\begin{equation} \label{}
\small
    O(Baseline) = O(1) + T.R_T. (O(clients) + O(server))
\end{equation}
\begin{equation} \label{}
\small
\begin{split}
    O(Baseline) &= O(1) + T.R_T. (L.O(1client) \\
                & + O(server))
\end{split}
\end{equation}
\begin{equation} \label{}
\small
    O(Baseline) = O(1) + T.R_T. (L.O(1client) + O(1))
\end{equation}
\begin{equation} \label{}
\small
\begin{split}
    O(Baseline) &= O(1) + T.R_T. L. O(E.\sum_{b=1}^{\beta} N_{bl}^t+O(1)) \\
                & + O(T.R_T.)
\end{split}
\end{equation}
\begin{equation} \label{}
\small
\begin{split}
    O(Baseline) &= O(1) + O(T.R_T. L.E.\sum_{b=1}^{\beta} N_{bl}^t) \\
                &+ O(T.R_T.L.) + O(T.R_T)
\end{split}
\end{equation}
\begin{equation} \label{}
\small
\begin{split}
    O(Baseline) &= O(T.R_T. L.E.\sum_{b=1}^{\beta} N_{bl}^t) \\
                &+ O(T.R_T. L.) + O(T.R_T)
\end{split}
\end{equation}
From the definition above that $\sum_{b=1}^{\beta} N_{bl}^t = N_l^t$, $R_T=R/T$, $N_l=T.N_l^t=T$, and $L \geq 1$ therefore the complexity of the baseline will be
\begin{equation} \label{}
\small
\begin{split}
    O(Baseline) &= O(T.R/T.L.E.N_{l}^t)+  O(T.R/T.L.) \\
                &+ O(T.R/T) 
\end{split}
\end{equation}
\begin{equation} \label{}
\small
    O(Baseline) = O(R.L.E.N_{l}^t)+ O(R.L) + O(R)
\end{equation}
\begin{equation} \label{}
\small
    O(Baseline) = O(R.L.E.N_{l}^t)
\end{equation}
Since $N_{l}^t < N_{l} $ and E is set as a small constant in our method i.e. 2, then the baseline complexity will be:
\begin{equation} \label{}
\small
    O(Baseline) = O(R.L.N_{l})
\end{equation}

\section{Complete Numerical Result of Ablation Study}\label{app:ablation}
In this section, we present the complete numerical result of the ablation study. Table \ref{tab:complete_ablation} shows the detailed numerical result of PIP with various configurations in CIFAR100 dataset with T=10.
\begin{table*}[]
\small
\centering
\begin{tabular}{ccccccccccccc}
\hline
Config. & Prompt & Proto+Aug & W.Agg & Head & 10             & 20             & 30             & 40             & 50             & Avg            & PD             & Imp        \\ \hline
A       & \checkmark      &       &       &      & 82.15          & 69.97          & 67.06          & 64.52          & 63.78          & -              & -              & -          \\
B       & \checkmark      & \checkmark     &       &      & 90.99          & 85.46          & 83.65          & 81.99          & 80.90          & -              & -              & -          \\
C       & \checkmark      &       & \checkmark     &      & 82.13          & 69.85          & 67.18          & 65.15          & 64.11          & -              & -              & -          \\
D       & \checkmark      & \checkmark     & \checkmark     &      & 91.37          & 86.46          & 84.37          & 82.79          & 81.87          & -              & -              & -          \\
E       & \checkmark      &       &       & \checkmark    & 96.60          & 83.20          & 77.10          & 74.08          & 70.86          & -              & -              & -          \\
F       & \checkmark      & \checkmark     &       & \checkmark    & 98.90          & 91.80          & 88.80          & 87.25          & 86.78          & -              & -              & -          \\
G       & \checkmark      &       & \checkmark     & \checkmark    & 98.60          & 71.80          & 68.73          & 67.68          & 68.44          & -              & -              & -          \\ 
PIP     & \checkmark      & \checkmark     & \checkmark     & \checkmark    & \textbf{98.60} & \textbf{92.90} & \textbf{89.57} & \textbf{87.78} & \textbf{87.14} & -              & -              & -          \\ \hline
Config. & Prompt & Proto+Aug & W.Agg & Head & 60             & 70             & 80             & 90             & 100            & Avg            & PD             & Imp        \\ \hline
A       & \checkmark      &       &       &      & 62.99          & 62.72          & 61.90          & 61.12          & 60.66          & 65.69          & 21.49          & 22.59      \\
B       & \checkmark      & \checkmark     &       &      & 79.07          & 78.63          & 78.99          & 78.69          & 77.56          & 81.59          & \textbf{13.43} & 6.69       \\
C       & \checkmark      &       & \checkmark     &      & 63.50          & 62.84          & 62.78          & 62.16          & 61.35          & 66.10          & 20.78          & 22.18      \\
D       & \checkmark      & \checkmark     & \checkmark     &      & 79.72          & 79.78          & 79.74          & 79.54          & 78.47          & 82.41          & \textbf{12.90} & 5.87       \\
E       & \checkmark      &       &       & \checkmark    & 68.73          & 71.83          & 70.00          & 70.71          & 70.68          & 75.38          & 25.92          & 12.90      \\
F       & \checkmark      & \checkmark     &       & \checkmark    & 85.52          & 85.29          & 84.83          & 84.86          & 84.17          & 87.82          & 14.73          & 0.46       \\
G       & \checkmark      &       & \checkmark     & \checkmark    & 69.48          & 70.31          & 70.98          & 69.77          & 69.94          & 72.57          & 28.66          & 15.71      \\
PIP     & \checkmark      & \checkmark     & \checkmark     & \checkmark    & \textbf{85.93} & \textbf{85.60} & \textbf{85.45} & \textbf{85.22} & \textbf{84.60} & \textbf{88.28} & \textbf{14.00} & \textbf{-} \\ \hline
\end{tabular}
\caption{Complete numerical result of ablation study on CIFAR100 dataset (T=10) on one seeded run i.e. 2021.  ”Avg” denotes the average accuracy of all tasks, ”PD” denotes performance drop, and ”Imp” denotes
improvement/gap of PIP compared to the respective configuration}
\label{tab:complete_ablation}
\end{table*}

\section{Complete Numerical Result of Experiment on 3 Benchmark Datasets with 
 T=5 and T=20 Settings}\label{app:task_size}
In this section, we present the detailed numerical results on different task sizes (T=5) and (T=20) on CIFAR100, MiniImageNet, and TinyImageNet datasets. Table \ref{tab:cifar_miniimagenet_t5} presents the complete numerical result on CIFAR100 and MiniImageNet dataset with T=5 setting, while the complete numerical result on TinyImageNet dataset with T=5 is shown in Table \ref{tab:tinymagenet_t5}. Table \ref{tab:cifar_t20}, \ref{tab:miniimagenet_t20}, and \ref{tab:tinyimagenet_t20} presents detailed numerical results with T=20 setting in CIFAR100, MiniImageNet, and TinyImageNet respectively.

\begin{table*}[h]
\small
\centering
\setlength{\tabcolsep}{4pt}
\begin{tabular}{l|cccccccc|cccccccc}
\hline
                   & \multicolumn{8}{c}{CIFAR100 (T=5)}   & \multicolumn{8}{c}{MiniImageNet (T=5)}  \\ \hline
Method             & 20    & 40    & 60    & 80    & 100   & Avg   & PD    & Imp   & 20     & 40    & 60    & 80    & 100   & Avg   & PD    & Imp   \\ \hline
iCaRL+FL           & 77.00          & 59.60          & 51.90          & 44.40          & 39.60          & 54.50          & 37.40          & 30.91          & 73.50          & 56.20          & 46.20          & 40.20          & 35.50          & 50.32          & 38.00         & 39.53 \\
BiC+FL             & 78.40          & 60.40          & 53.20          & 47.50          & 41.20          & 56.14          & 37.20          & 29.27          & 72.60          & 56.80          & 49.20          & 43.50          & 38.70          & 52.16          & 33.90         & 37.69 \\
PODNet+FL          & 77.60          & 62.10          & 56.30          & 50.80          & 43.30          & 58.02          & 34.30          & 27.39          & 73.10          & 58.40          & 53.20          & 46.50          & 43.40          & 54.92          & 29.70         & 34.93 \\
DDE+iCaRL+FL       & 77.00          & 60.20          & 55.70          & 49.30          & 42.50          & 56.94          & 34.50          & 28.47          & 72.30          & 57.20          & 51.70          & 44.30          & 41.30          & 53.36          & 31.00         & 36.49 \\
GeoDL+iCaRL+FL     & 72.50          & 61.10          & 54.00          & 49.50          & 44.50          & 56.32          & 28.00          & 29.09          & 71.80          & 59.60          & 52.30          & 46.10          & 42.50          & 54.46          & 29.30         & 35.39 \\
SS-IL+FL           & 78.10          & 61.80          & 52.80          & 48.80          & 46.00          & 57.50          & 32.10          & 27.91          & 66.50          & 52.10          & 42.60          & 36.70          & 36.50          & 46.88          & 30.00         & 42.97 \\
DyTox+FL           & 78.80          & 70.50          & 63.90          & 59.90          & 55.90          & 65.80          & 22.90          & 19.61          & 69.60          & 64.20          & 59.10          & 53.40          & 48.50          & 58.96          & 21.10         & 30.89 \\
AFC+FL             & 71.10          & 63.80          & 58.40          & 53.60          & 46.40          & 58.66          & 24.70          & 26.75          & 78.00          & 64.50          & 57.00          & 51.30          & 47.30          & 59.62          & 30.70         & 30.23 \\
GLFC               & 83.70          & 75.50          & 66.50          & 62.10          & 53.80          & 68.32          & 29.90          & 17.09          & 79.70          & 73.40          & 65.20          & 58.10          & 51.80          & 65.64          & 27.90         & 24.21 \\
LGA                & 83.30          & 77.30          & 72.80          & 67.80          & 63.70          & 72.98          & 19.60          & 12.43          & 78.90          & 75.50          & 68.10          & 62.10          & 61.90          & 69.30          & 17.00         & 20.55 \\
TARGET             & 77.10          & 39.08          & 26.95          & 20.71          & 17.72          & 36.31          & 59.38          & 49.10          & 61.75          & 28.48          & 18.33          & 11.95          & 7.58           & 25.62          & 54.17         & 64.23 \\ \hline
Fed-L2P            & 89.50          & 69.70          & 63.33          & 59.89          & 58.35          & 68.15          & 31.15          & 17.26          & 91.40          & 77.38          & 74.03          & 72.73          & 74.91          & 78.09          & 16.49         & 11.76 \\
Fed-DualP          & 93.15          & 76.88          & 67.55          & 66.56          & 66.06          & 74.04          & 27.09          & 11.37          & 94.35          & 81.13          & 74.67          & 73.51          & 76.67          & 80.06          & 17.68         & 9.79  \\
\textbf{PIP-L2P}   & 92.55          & \textbf{87.10} & \textbf{84.40} & \textbf{82.85} & \textbf{81.41} & \textbf{85.66} & \textbf{11.14} & \textbf{-0.25} & 93.70          & \textbf{89.83} & \textbf{88.97} & \textbf{88.45} & 86.89          & 89.57          & \textbf{6.81} & 0.28  \\
\textbf{PIP-DualP} & \textbf{96.75} & 86.35          & 83.57          & 80.68          & 79.71          & 85.41          & 17.04          & -              & \textbf{95.30} & 89.68          & 88.52          & 88.44          & \textbf{87.32} & \textbf{89.85} & 7.98          & -    \\ \hline

\end{tabular}
\caption{Complete numerical results on CIFAR100 and MniImageNet dataset (T=5) in one-seeded run i.e. 2021}
\label{tab:cifar_miniimagenet_t5}
\end{table*}

\begin{table*}[h]
\small
\centering
\begin{tabular}{lcccccccc}
\hline
                   & \multicolumn{8}{c}{TinyImageNet (T=5)}                                                                                               \\ \hline
Method             & 40             & 80             & 120            & 160            & 200            & Avg            & PD             & Imp           \\ \hline
iCaRL+FL           & 65.00          & 48.00          & 42.70          & 38.70          & 35.00          & 45.88          & 30.00          & 38.31         \\
BiC+FL             & 65.70          & 48.70          & 43.00          & 40.30          & 35.70          & 46.68          & 30.00          & 37.51         \\
PODNet+FL          & 66.00          & 50.30          & 44.70          & 41.30          & 37.00          & 47.86          & 29.00          & 36.33         \\
DDE+iCaRL+FL       & 63.00          & 51.30          & 45.30          & 41.00          & 36.00          & 47.32          & 27.00          & 36.87         \\
GeoDL+iCaRL+FL     & 65.30          & 50.00          & 45.00          & 40.70          & 36.00          & 47.40          & 29.30          & 36.79         \\
SS-IL+FL           & 65.00          & 42.30          & 38.30          & 35.00          & 30.30          & 42.18          & 34.70          & 42.01         \\
DyTox+FL           & 58.60          & 43.10          & 41.60          & 37.20          & 32.90          & 42.68          & 25.70          & 41.51         \\
AFC+FL             & 62.50          & 52.10          & 45.70          & 43.20          & 35.70          & 47.84          & 26.80          & 36.35         \\
GLFC               & 66.00          & 55.30          & 49.00          & 45.00          & 40.30          & 51.12          & 25.70          & 33.07         \\
LGA                & 67.70          & 59.80          & 53.50          & 47.90          & 43.80          & 54.54          & 23.90          & 29.65         \\
TARGET             & 53.25          & 27.08          & 17.77          & 11.04          & 7.28           & 23.28          & 45.97          & 60.91         \\ \hline
Fed-L2P            & 82.70          & 70.23          & 66.33          & 62.51          & 63.73          & 69.10          & 18.97          & 15.09         \\
Fed-DualP          & 85.10          & 67.23          & 60.25          & 55.99          & 58.59          & 65.43          & 26.51          & 18.76         \\
\textbf{PIP-L2P}   & 82.70          & 70.23          & 66.33          & 62.51          & 63.73          & 69.10          & 18.97          & 15.09         \\ 
\textbf{PIP-DualP} & \textbf{91.20} & \textbf{84.95} & \textbf{83.80} & \textbf{81.86} & \textbf{79.16} & \textbf{84.19} & \textbf{12.04} & \textbf{0.00} \\ \hline
\end{tabular}
\caption{Complete numerical results on TinyImageNet dataset (T=5) in one seeded run i.e. 2021}
\label{tab:tinymagenet_t5}
\end{table*}

\begin{table*}[]
\renewcommand{\arraystretch}{0.90}
\small
\footnotesize
% \scriptsize
\centering
\begin{tabular}{lccccccccccccc}
\hline
                   & \multicolumn{13}{c}{CIFAR100 (T=20)} \\ \hline
Method             & 5              & 10             & 15             & 20             & 25             & 30             & 35             & 40             & 45             & 50             & Avg            & PD             & Imp        \\ \hline
iCaRL+FL           & 82.00          & 80.00          & 67.00          & 62.00          & 61.30          & 60.30          & 57.00          & 54.30          & 53.00          & 51.70          & -              & -              & -          \\
BiC+FL             & 82.00          & 77.30          & 68.30          & 64.00          & 63.70          & 62.30          & 60.30          & 58.70          & 55.00          & 53.30          & -              & -              & -          \\
PODNet+FL          & 83.00          & 76.30          & 70.30          & 68.00          & 66.30          & 67.00          & 65.30          & 61.70          & 61.30          & 58.70          & -              & -              & -          \\
DDE+iCaRL+FL       & 83.00          & 75.30          & 69.70          & 65.00          & 67.00          & 63.70          & 59.30          & 58.00          & 60.00          & 55.30          & -              & -              & -          \\
GeoDL+iCaRL+FL     & 82.00          & 78.30          & 71.30          & 67.70          & 68.00          & 65.30          & 64.30          & 60.00          & 58.70          & 56.00          & -              & -              & -          \\
SS-IL+FL           & 83.00          & 73.30          & 63.70          & 61.30          & 60.30          & 59.30          & 57.30          & 56.00          & 54.70          & 53.30          & -              & -              & -          \\
DyTox+FL           & 79.60          & 78.30          & 67.10          & 65.60          & 68.50          & 64.30          & 63.70          & 61.00          & 58.80          & 59.00          & -              & -              & -          \\
AFC+FL             & 75.60          & 69.60          & 57.10          & 58.50          & 45.50          & 55.40          & 51.40          & 50.40          & 45.20          & 42.40          & -              & -              & -          \\
GLFC               & 82.20          & 82.50          & 74.90          & 75.20          & 73.30          & 71.50          & 70.10          & 67.70          & 64.60          & 65.90          & -              & -              & -          \\
LGA                & 85.80          & 85.90          & 80.70          & 78.90          & 78.40          & 74.60          & 75.10          & 71.30          & 68.90          & 69.20          & -              & -              & -          \\
TARGET             & 34.20          & 15.70          & 9.93           & 8.60           & 4.64           & 3.90           & 3.23           & 2.72           & 2.56           & 1.86           & -              & -              & -          \\ \hline
Fed-L2P            & 94.00          & 87.70          & 74.13          & 74.70          & 74.36          & 71.90          & 74.06          & 72.40          & 71.71          & 68.96          & -              & -              & -          \\ 
Fed-DualP          & 94.80          & 91.40          & 85.13          & 85.00          & 84.20          & 80.17          & 79.14          & 76.75          & 76.60          & 75.46          & -              & -              & -          \\
\textbf{PIP-L2P}   & 94.60          & 88.70          & 80.07          & 76.85          & 72.76          & 69.83          & 68.97          & 67.97          & 67.91          & 65.82          & -              & -              & -          \\ 
\textbf{PIP-DualP} & \textbf{98.40} & \textbf{96.70} & \textbf{92.67} & \textbf{91.65} & \textbf{90.88} & \textbf{88.47} & \textbf{88.37} & \textbf{86.87} & \textbf{87.00} & \textbf{86.30} & \textbf{-}     & \textbf{-}     & \textbf{-} \\ \hline
Method             & 55             & 60             & 65             & 70             & 75             & 80             & 85             & 90             & 95             & 100            & Avg            & PD             & Imp        \\ \hline    
iCaRL+FL           & 50.30          & 50.00          & 48.70          & 48.00          & 46.70          & 45.00          & 45.00          & 44.00          & 43.30          & 42.70          & 44.00          & 39.30          & 38.38      \\
BiC+FL             & 52.00          & 51.30          & 50.30          & 49.70          & 48.00          & 47.00          & 46.30          & 45.70          & 45.30          & 44.30          & 45.72          & 37.70          & 36.66      \\
PODNet+FL          & 56.30          & 55.00          & 54.00          & 53.00          & 51.00          & 50.30          & 49.30          & 48.00          & 48.30          & 47.70          & 48.72          & 35.30          & 33.66      \\
DDE+iCaRL+FL       & 54.70          & 54.00          & 53.30          & 52.00          & 50.70          & 50.00          & 49.30          & 48.70          & 48.00          & 47.30          & 48.66          & 35.70          & 33.72      \\
GeoDL+iCaRL+FL     & 55.30          & 55.00          & 53.70          & 53.00          & 51.70          & 50.70          & 50.00          & 49.00          & 49.30          & 48.00          & 49.40          & 34.00          & 32.98      \\
SS-IL+FL           & 52.30          & 52.00          & 51.30          & 50.70          & 50.00          & 49.30          & 49.00          & 48.30          & 48.00          & 47.70          & 48.46          & 35.30          & 33.92      \\
DyTox+FL           & 56.20          & 58.50          & 58.30          & 58.20          & 55.00          & 51.80          & 49.70          & 48.70          & 49.00          & 52.70          & 50.38          & 26.90          & 32.00      \\
AFC+FL             & 41.30          & 35.60          & 37.10          & 37.80          & 38.90          & 35.20          & 34.40          & 34.50          & 36.20          & 33.80          & 34.82          & 41.80          & 47.56      \\
GLFC               & 63.70          & 64.20          & 62.00          & 61.00          & 60.20          & 58.90          & 57.60          & 59.30          & 56.80          & 56.80          & 57.88          & 25.40          & 24.50      \\
LGA                & 68.30          & 67.70          & 65.50          & 65.60          & 64.00          & 63.00          & 63.10          & 63.70          & 61.60          & 60.50          & 62.38          & 25.30          & 20.00      \\
TARGET             & 1.45           & 1.03           & 1.42           & 1.57           & 1.71           & 1.38           & 1.67           & 2.13           & 2.19           & 1.94           & 1.86           & 32.26          & 80.52      \\ \hline
Fed-L2P            & 68.95          & 66.90          & 65.52          & 66.56          & 64.68          & 65.20          & 63.56          & 64.81          & 63.00          & 62.46          & 63.81          & 31.54          & 18.57      \\
Fed-DualP          & 75.60          & 72.80          & 71.92          & 72.37          & 70.89          & 70.94          & 71.16          & 71.11          & 70.89          & 70.30          & 70.88          & 24.50          & 11.50      \\
\textbf{PIP-L2P}   & 63.02          & 60.37          & 59.14          & 58.21          & 59.39          & 59.51          & 62.15          & 63.06          & 61.72          & 60.33          & 61.35          & 34.27          & 21.03      \\
\textbf{PIP-DualP} & \textbf{84.53} & \textbf{83.70} & \textbf{82.88} & \textbf{83.16} & \textbf{82.08} & \textbf{82.54} & \textbf{82.69} & \textbf{82.39} & \textbf{82.78} & \textbf{81.49} & \textbf{82.38} & \textbf{16.91} & \textbf{-} \\ \hline
\end{tabular}
\caption{Complete numerical results on CIFAR100 dataset (T=20) in one-seeded run i.e. 2021}
\label{tab:cifar_t20}
\end{table*}

\begin{table*}[]
\renewcommand{\arraystretch}{0.90}
% \footnotesize
\small
\centering
\begin{tabular}{lccccccccccccc}
\hline
                   & \multicolumn{13}{c}{MiniImageNet (T=20)} \\ \hline
Method             & 5              & 10             & 15             & 20             & 25             & 30             & 35             & 40             & 45             & 50             & Avg            & PD            & Imp        \\ \hline
iCaRL+FL           & 83.00          & 66.00          & 61.30          & 56.00          & 56.30          & 53.00          & 49.70          & 47.00          & 46.30          & 46.00          & -              & -             & -          \\
BiC+FL             & 82.30          & 64.70          & 59.00          & 58.30          & 57.00          & 54.70          & 52.30          & 50.30          & 49.00          & 47.70          & -              & -             & -          \\
PODNet+FL          & 81.70          & 63.30          & 60.30          & 59.30          & 58.30          & 56.30          & 55.00          & 53.30          & 51.70          & 50.00          & -              & -             & -          \\
DDE+iCaRL+FL       & 80.00          & 60.70          & 58.70          & 56.30          & 57.00          & 55.30          & 53.00          & 51.70          & 50.30          & 49.30          & -              & -             & -          \\
GeoDL+iCaRL+FL     & 82.30          & 66.30          & 62.70          & 61.00          & 60.30          & 58.00          & 56.30          & 55.30          & 53.00          & 51.30          & -              & -             & -          \\
SS-IL+FL           & 80.00          & 65.30          & 61.70          & 57.30          & 56.30          & 54.00          & 51.30          & 50.00          & 49.30          & 48.30          & -              & -             & -          \\
DyTox+FL           & 71.60          & 52.70          & 61.60          & 53.20          & 56.80          & 48.90          & 45.70          & 49.40          & 39.10          & 44.10          & -              & -             & -          \\
AFC+FL             & 72.40          & 53.00          & 51.80          & 38.10          & 41.40          & 39.60          & 41.20          & 37.20          & 33.70          & 32.40          & -              & -             & -          \\
GLFC               & 84.00          & 71.70          & 70.00          & 69.30          & 67.30          & 66.30          & 61.00          & 60.70          & 59.30          & 58.70          & -              & -             & -          \\
LGA                & 78.80          & 79.40          & 76.80          & 73.50          & 69.80          & 68.50          & 67.30          & 66.10          & 63.80          & 62.10          & -              & -             & -          \\
TARGET             & 41.40          & 12.00          & 6.67           & 5.00           & 4.00           & 4.77           & 2.86           & 2.50           & 2.22           & 2.00           & -              & -             & -          \\ \hline
Fed-L2P            & 94.00          & 92.50          & 87.47          & 86.05          & 83.52          & 83.60          & 81.91          & 81.13          & 81.87          & 81.20          & -              & -             & -          \\
Fed-DualP          & 96.20          & 94.50          & 88.73          & 86.80          & 86.04          & 86.80          & 85.83          & 85.63          & 86.11          & 85.90          & -              & -             & -          \\
\textbf{PIP-L2P}   & 94.20          & 92.10          & 90.80          & 89.05          & 86.92          & 86.77          & 85.49          & 84.20          & 85.84          & 83.94          & -              & -             & -          \\
\textbf{PIP-DualP} & \textbf{97.80} & \textbf{96.70} & \textbf{94.33} & \textbf{92.45} & \textbf{91.92} & \textbf{91.70} & \textbf{91.09} & \textbf{90.80} & \textbf{91.07} & \textbf{90.34} & \textbf{-}     & \textbf{-}    & \textbf{-} \\ \hline
Method             & 55             & 60             & 65             & 70             & 75             & 80             & 85             & 90             & 95             & 100            & Avg            & PD            & Imp        \\ \hline
iCaRL+FL           & 44.00          & 42.30          & 40.00          & 39.70          & 37.30          & 36.00          & 34.70          & 34.30          & 33.00          & 32.00          & 34.00          & 51.00         & 54.17      \\
BiC+FL             & 46.70          & 44.00          & 42.70          & 41.30          & 40.30          & 38.00          & 37.00          & 36.30          & 34.70          & 33.00          & 35.80          & 49.30         & 52.37      \\
PODNet+FL          & 49.30          & 48.00          & 47.00          & 45.30          & 44.70          & 43.70          & 42.00          & 39.70          & 38.70          & 37.00          & 40.22          & 44.70         & 47.95      \\
DDE+iCaRL+FL       & 48.70          & 48.30          & 47.70          & 46.70          & 45.70          & 44.30          & 42.30          & 40.00          & 38.30          & 37.30          & 40.44          & 42.70         & 47.73      \\
GeoDL+iCaRL+FL     & 50.00          & 48.70          & 48.00          & 46.30          & 45.00          & 44.00          & 41.70          & 40.00          & 38.00          & 36.70          & 40.08          & 45.60         & 48.09      \\
SS-IL+FL           & 47.00          & 45.00          & 44.30          & 43.00          & 41.30          & 40.70          & 39.30          & 38.70          & 37.00          & 36.00          & 38.34          & 44.00         & 49.83      \\
DyTox+FL           & 37.70          & 35.20          & 33.60          & 31.50          & 28.60          & 27.30          & 27.10          & 26.50          & 25.80          & 24.90          & 26.32          & 46.70         & 61.85      \\
AFC+FL             & 29.70          & 33.50          & 29.60          & 30.20          & 25.10          & 25.10          & 26.10          & 24.60          & 24.00          & 23.50          & 24.66          & 48.90         & 63.51      \\
GLFC               & 55.30          & 53.00          & 52.00          & 50.30          & 49.70          & 47.30          & 46.00          & 42.70          & 40.30          & 39.00          & 43.06          & 45.00         & 45.11      \\
LGA                & 60.60          & 59.80          & 57.20          & 56.80          & 55.10          & 54.70          & 54.10          & 53.20          & 51.60          & 48.20          & 52.36          & 30.60         & 35.81      \\
TARGET             & 1.82           & 1.67           & 1.54           & 1.43           & 1.33           & 1.25           & 1.18           & 1.11           & 1.05           & 1.00           & 1.12           & 40.40         & 87.05      \\ \hline
Fed-L2P            & 80.58          & 81.08          & 81.15          & 79.96          & 78.88          & 80.41          & 78.84          & 77.81          & 78.80          & 78.68          & 78.91          & 15.32         & 9.26       \\
Fed-DualP          & 85.82          & 85.97          & 85.78          & 84.63          & 84.71          & 84.53          & 82.81          & 81.73          & 81.96          & 82.35          & 82.68          & 13.85         & 5.49       \\
\textbf{PIP-L2P}   & 83.16          & 82.95          & 82.86          & 81.56          & 79.69          & 79.91          & 79.59          & 78.14          & 80.23          & 78.85          & 79.35          & 15.35         & 8.82       \\
\textbf{PIP-DualP} & \textbf{89.69} & \textbf{90.03} & \textbf{89.97} & \textbf{89.43} & \textbf{89.20} & \textbf{89.14} & \textbf{88.38} & \textbf{87.39} & \textbf{87.89} & \textbf{88.04} & \textbf{88.17} & \textbf{9.76} & \textbf{-}
\\ \hline
\end{tabular}
\caption{Complete numerical results on MiniImageNet dataset (T=20) in one-seeded run i.e. 2021}
\label{tab:miniimagenet_t20}
\end{table*}

\begin{table*}[]
\footnotesize
\centering
\begin{tabular}{lccccccccccccc}
\hline
                   & \multicolumn{13}{c}{TinyImageNet (T=20)}  \\ \hline
Method             & 10             & 20             & 30             & 40             & 50             & 60             & 70             & 80             & 90             & 100            & Avg            & PD            & Imp        \\ \hline
iCaRL+FL           & 67.00          & 59.30          & 54.00          & 48.30          & 46.70          & 44.70          & 43.30          & 39.00          & 37.30          & 33.00          & -              & -             & -          \\
BiC+FL             & 67.30          & 59.70          & 54.70          & 50.00          & 48.30          & 45.30          & 43.00          & 40.70          & 38.00          & 33.70          & -              & -             & -          \\
PODNet+FL          & 69.00          & 59.30          & 55.00          & 51.70          & 50.00          & 46.70          & 43.70          & 41.00          & 39.30          & 38.00          & -              & -             & -          \\
DDE+iCaRL+FL       & 70.00          & 59.30          & 53.30          & 51.00          & 48.30          & 45.70          & 42.30          & 40.00          & 38.00          & 36.30          & -              & -             & -          \\
GeoDL+iCaRL+FL     & 66.30          & 56.70          & 51.00          & 49.70          & 44.70          & 42.30          & 41.00          & 39.00          & 37.30          & 35.00          & -              & -             & -          \\
SS-IL+FL           & 66.70          & 54.00          & 47.70          & 45.30          & 42.30          & 42.00          & 40.70          & 38.00          & 36.00          & 34.30          & -              & -             & -          \\
DyTox+FL           & 77.60          & 70.20          & 63.40          & 56.60          & 52.00          & 44.60          & 51.60          & 39.60          & 41.50          & 39.00          & -              & -             & -          \\
AFC+FL             & 74.00          & 62.90          & 57.60          & 54.20          & 45.10          & 44.40          & 40.70          & 36.90          & 33.00          & 33.60          & -              & -             & -          \\
GLFC               & 68.70          & 63.30          & 61.70          & 57.30          & 56.00          & 53.00          & 50.30          & 47.70          & 46.30          & 45.00          & -              & -             & -          \\
LGA                & 74.00          & 67.60          & 64.90          & 61.00          & 58.90          & 55.70          & 53.60          & 51.30          & 50.10          & 48.80          & -              & -             & -          \\
TARGET             & 28.00          & 12.40          & 7.67           & 7.40           & 5.60           & 3.67           & 2.63           & 2.12           & 1.93           & 1.88           & -              & -             & -          \\ \hline
Fed-L2P            & 75.60          & 72.50          & 74.80          & 74.60          & 72.64          & 74.97          & 74.66          & 74.70          & 73.49          & 70.94          & -              & -             & -          \\
Fed-DualP          & 79.40          & 78.90          & 79.00          & 78.10          & 76.48          & 77.53          & 77.14          & 76.85          & 76.04          & 75.48          & -              & -             & -          \\
\textbf{PIP-L2P}   & 79.20          & 78.80          & 81.07          & 79.50          & 76.60          & 79.80          & 77.00          & 78.23          & 76.09          & 75.80          & -              & -             & -          \\
\textbf{PIP-DualP} & \textbf{89.20} & \textbf{87.80} & \textbf{86.93} & \textbf{85.00} & \textbf{84.12} & \textbf{85.70} & \textbf{86.31} & \textbf{86.62} & \textbf{86.76} & \textbf{86.44} & \textbf{-}     & \textbf{-}    & \textbf{-} \\ \hline
Method             & 110            & 120            & 130            & 140            & 150            & 160            & 170            & 180            & 190            & 200            & Avg            & PD            & Imp        \\ \hline
iCaRL+FL           & 32.00          & 30.30          & 28.00          & 27.00          & 26.30          & 25.30          & 24.70          & 24.00          & 22.70          & 22.00          & 23.74          & 45.00         & 60.10      \\
BiC+FL             & 32.70          & 32.30          & 30.30          & 29.00          & 27.70          & 27.30          & 26.00          & 25.70          & 24.30          & 23.30          & 25.32          & 44.00         & 58.52      \\
PODNet+FL          & 37.00          & 35.70          & 34.70          & 34.00          & 33.00          & 32.30          & 31.00          & 30.00          & 29.30          & 28.00          & 30.12          & 41.00         & 53.72      \\
DDE+iCaRL+FL       & 35.00          & 33.70          & 32.00          & 31.00          & 30.30          & 30.00          & 28.70          & 28.30          & 27.30          & 26.00          & 28.06          & 44.00         & 55.78      \\
GeoDL+iCaRL+FL     & 33.70          & 32.00          & 31.00          & 30.30          & 28.70          & 28.00          & 27.30          & 26.30          & 25.00          & 24.70          & 26.26          & 41.60         & 57.58      \\
SS-IL+FL           & 33.00          & 31.00          & 29.30          & 28.30          & 27.70          & 27.00          & 26.30          & 26.00          & 25.00          & 24.30          & 25.72          & 42.40         & 58.12      \\
DyTox+FL           & 37.80          & 31.20          & 34.20          & 30.60          & 29.80          & 29.20          & 28.30          & 27.50          & 26.80          & 15.30          & 25.42          & 62.30         & 58.42      \\
AFC+FL             & 30.80          & 28.90          & 27.10          & 22.80          & 24.50          & 23.60          & 22.10          & 20.70          & 18.40          & 18.10          & 20.58          & 55.90         & 63.26      \\
GLFC               & 42.70          & 41.00          & 40.00          & 39.30          & 38.00          & 36.70          & 35.30          & 34.00          & 33.00          & 31.70          & 34.14          & 37.00         & 49.70      \\
LGA                & 45.20          & 43.70          & 42.80          & 41.20          & 40.50          & 38.90          & 37.40          & 36.60          & 35.10          & 33.80          & 36.36          & 40.20         & 47.48      \\
TARGET             & 1.69           & 1.48           & 1.26           & 1.27           & 1.17           & 1.02           & 0.66           & 0.60           & 0.53           & 0.50           & 0.66           & 27.50         & 83.18      \\ \hline
Fed-L2P            & 71.35          & 70.38          & 70.46          & 67.47          & 67.20          & 65.20          & 66.16          & 65.23          & 68.61          & 65.94          & 66.23          & 9.66          & 17.61      \\
Fed-DualP          & 75.20          & 75.37          & 75.23          & 73.90          & 73.76          & 72.80          & 73.27          & 72.20          & 72.28          & 71.69          & 72.45          & 7.71          & 11.39      \\
\textbf{PIP-L2P}   & 77.65          & 77.12          & 75.86          & 75.29          & 75.57          & 73.66          & 73.22          & 71.48          & 72.13          & 70.15          & 72.13          & 9.05          & 11.71      \\
\textbf{PIP-DualP} & \textbf{86.80} & \textbf{86.22} & \textbf{86.22} & \textbf{85.67} & \textbf{85.81} & \textbf{84.30} & \textbf{84.35} & \textbf{83.49} & \textbf{83.72} & \textbf{83.33} & \textbf{83.84} & \textbf{5.87} & \textbf{-} \\ \hline
\end{tabular}
\caption{Complete numerical results on TinyImageNet dataset (T=20) in one-seeded run i.e. 2021}
\label{tab:tinyimagenet_t20}
\end{table*}

\vspace{-5pt}
\section{Complete Numerical Result of Experiment on Smaller Local Clients}\label{app:smaller_nclients}
\vspace{-3pt}
In this section, we present the detailed numerical results of smaller local clients on CIFAR100, and TinyImageNet datasets.  Table \ref{tab:cifar_nclients}, and \ref{tab:tinyimagenet_nclients} presents detailed numerical results in CIFAR100, and TinyImageNet respectively with various numbers of selected local clients.

\begin{table*}[h]
\footnotesize
\setlength{\tabcolsep}{4pt}
\centering
\begin{tabular}{lcccccccccccccc}
\hline
          &                     & \multicolumn{13}{c}{CIFAR100 (T=10)} \\ \hline
Method    & \#Local Clients (L) & 10             & 20             & 30             & 40             & 50             & 60             & 70             & 80             & 90             & 100            & AVG            & PD             & Imp           \\ \hline
GLFC      & 10 (33.33\%)        & 90.00          & 82.30          & 77.00          & 72.30          & 65.00          & 66.30          & 59.70          & 56.30          & 50.30          & 50.00          & 66.92          & 40.00          & 21.36         \\
LGA       & 10 (33.33\%)        & 89.60          & 83.20          & 79.30          & 76.10          & 72.90          & 71.70          & 68.40          & 65.70          & 64.70          & 62.90          & 73.45          & 26.70          & 14.83         \\
TARGET    & 10 (33.33\%)        & 65.10          & 37.15          & 26.50          & 21.25          & 19.00          & 16.43          & 14.40          & 13.19          & 11.38          & 10.03          & 23.44          & 55.07          & 64.84         \\
Fed-L2P   & 10 (33.33\%)        & 94.60          & 74.70          & 70.10          & 68.88          & 65.34          & 66.30          & 67.59          & 67.00          & 67.03          & 67.11          & 70.86          & 27.49          & 17.42         \\
Fed-DualP & 10 (33.33\%)        & 96.60          & 83.20          & 77.10          & 74.08          & 70.86          & 68.73          & 71.83          & 70.00          & 70.71          & 70.68          & 75.38          & 25.92          & 12.90         \\
PIP-L2P   & 10 (33.33\%)        & 95.30          & 90.05          & 85.97          & 83.35          & 81.82          & 79.82          & 79.69          & 78.68          & 79.01          & 77.28          & 83.10          & 18.02          & 5.18          \\
PIP-DualP & 10 (33.33\%)        & \textbf{98.60} & \textbf{92.90} & \textbf{89.57} & \textbf{87.78} & \textbf{87.14} & \textbf{85.93} & \textbf{85.60} & \textbf{85.45} & \textbf{85.22} & \textbf{84.60} & \textbf{88.28} & \textbf{14.00} & \textbf{0.00} \\ \hline
GLFC      & 3 (10.00\%)         & 86.70          & 74.50          & 71.70          & 65.63          & 65.26          & 60.37          & 54.51          & 52.63          & 48.30          & 44.56          & 62.42          & 42.14          & 24.64         \\
LGA       & 3 (10.00\%)         & 85.50          & 74.30          & 75.70          & 72.80          & 71.02          & 66.50          & 64.57          & 61.04          & 61.22          & 56.05          & 68.87          & 29.45          & 18.19         \\
TARGET    & 3 (10.00\%)         & 46.00          & 26.45          & 16.87          & 13.02          & 10.18          & 7.90           & 7.04           & 5.88           & 4.63           & 3.97           & 14.19          & 42.03          & 72.87         \\
Fed-L2P   & 3 (10.00\%)         & 80.40          & 74.20          & 73.77          & 71.00          & 71.30          & 69.65          & 68.29          & 69.04          & 69.44          & 72.33          & 71.94          & 8.07           & 15.12         \\
Fed-DualP & 3 (10.00\%)         & 88.30          & 83.20          & 80.27          & 77.45          & 74.26          & 71.73          & 73.91          & 73.34          & 72.42          & 74.72          & 76.96          & \textbf{13.58} & 10.10         \\ 
PIP-L2P   & 3 (10.00\%)         & 94.50          & 84.80          & 80.30          & 78.30          & 73.70          & 70.47          & 67.34          & 66.60          & 67.18          & 67.62          & 75.08          & 26.88          & 11.98         \\
PIP-DualP & 3 (10.00\%)         & \textbf{98.40} & \textbf{92.00} & \textbf{90.57} & \textbf{89.03} & \textbf{87.72} & \textbf{83.30} & \textbf{83.69} & \textbf{82.71} & \textbf{82.36} & \textbf{80.88} & \textbf{87.06} & 17.52          & \textbf{0.00} \\ \hline
GLFC      & 2 (6.67\%)          & 85.00          & 71.80          & 65.10          & 64.58          & 54.96          & 51.00          & 44.79          & 49.69          & 41.42          & 45.61          & 57.39          & 39.39          & 26.39         \\
LGA       & 2 (6.67\%)          & 86.40          & 76.35          & 78.07          & 67.75          & 63.40          & 62.67          & 61.57          & 56.61          & 60.39          & 53.61          & 66.68          & 32.79          & 17.10         \\
TARGET    & 2 (6.67\%)          & 49.70          & 24.60          & 16.67          & 12.45          & 9.80           & 7.53           & 6.74           & 5.41           & 4.69           & 4.20           & 14.18          & 45.50          & 69.60         \\
Fed-L2P   & 2 (6.67\%)          & 86.60          & 72.55          & 68.33          & 69.15          & 67.28          & 69.77          & 68.89          & 68.23          & 69.34          & 70.72          & 71.09          & \textbf{15.88} & 12.69         \\
Fed-DualP & 2 (6.67\%)          & 90.10          & 76.95          & 71.93          & 71.40          & 68.14          & 69.55          & 69.14          & 67.66          & 67.93          & 68.92          & 72.17          & 21.18          & 11.61         \\
PIP-L2P   & 2 (6.67\%)          & 83.10          & 66.40          & 69.10          & 70.30          & 64.76          & 63.23          & 59.99          & 56.58          & 57.52          & 55.86          & 64.68          & 27.24          & 19.10         \\
PIP-DualP & 2 (6.67\%)          & \textbf{96.70} & \textbf{90.15} & \textbf{88.80} & \textbf{86.70} & \textbf{83.34} & \textbf{79.20} & \textbf{79.34} & \textbf{78.30} & \textbf{77.94} & \textbf{77.28} & \textbf{83.78} & 19.42          & \textbf{0.00} \\ \hline
\end{tabular}
\caption{Complete numerical results on CIFAR100 (T=10) with smaller participating clients in one-seeded run i.e. 2021}
\label{tab:cifar_nclients}
\end{table*}

\begin{table*}[h]
\footnotesize
\centering
\setlength{\tabcolsep}{4pt}
\begin{tabular}{lcccccccccccccc}
\hline
          &                     & \multicolumn{13}{c}{TinyImageNet (T=10)}                                                                                                                                                                                   \\ \hline
Method    & \#Local Clients (L) & 20             & 40             & 60             & 80             & 100            & 120            & 140            & 160            & 180            & 200            & AVG            & PD              & Imp           \\ \hline
GLFC      & 10 (33.33\%)        & 66.00          & 58.30          & 55.30          & 51.00          & 47.70          & 45.30          & 43.00          & 40.00          & 37.30          & 35.00          & 47.89          & 31.00           & 38.90         \\
LGA       & 10 (33.33\%)        & 70.30          & 64.00          & 60.30          & 58.00          & 55.80          & 53.10          & 47.90          & 45.30          & 39.80          & 37.30          & 53.18          & 33.00           & 33.61         \\
TARGET    & 10 (33.33\%)        & 56.30          & 33.20          & 21.00          & 17.18          & 11.88          & 7.77           & 7.01           & 4.94           & 2.99           & 2.02           & 16.43          & 54.28           & 70.36         \\
Fed-L2P   & 10 (33.33\%)        & 80.80          & 74.40          & 72.53          & 70.53          & 67.24          & 68.62          & 67.99          & 67.93          & 65.70          & 64.50          & 70.02          & 16.30           & 16.77         \\
Fed-DualP & 10 (33.33\%)        & 86.60          & 75.70          & 71.33          & 66.33          & 64.28          & 62.83          & 62.27          & 61.19          & 59.99          & 62.60          & 67.31          & 24.00           & 19.48         \\
PIP-L2P   & 10 (33.33\%)        & 85.40          & 84.55          & 83.57          & 84.73          & 83.66          & 84.30          & 82.90          & 81.61          & 80.31          & 77.51          & 82.85          & \textbf{7.89}   & 3.94          \\
PIP-DualP & 10 (33.33\%)        & \textbf{92.70} & \textbf{87.90} & \textbf{86.83} & \textbf{88.23} & \textbf{87.04} & \textbf{87.18} & \textbf{85.96} & \textbf{85.13} & \textbf{84.62} & \textbf{82.35} & \textbf{86.79} & 10.35           & \textbf{0.00} \\ \hline
GLFC      & 3 (10.00\%)         & 55.60          & 37.70          & 26.37          & 19.88          & -              & -              & -              & -              & -              & -              & 34.89          & 35.73           & 50.11         \\
LGA       & 3 (10.00\%)         & 73.30          & 69.55          & 60.30          & 52.75          & 47.16          & -              & -              & -              & -              & -              & 60.61          & 26.14           & 24.39         \\
TARGET    & 3 (10.00\%)         & 36.70          & 20.20          & 12.37          & 8.60           & 5.06           & 3.53           & 2.89           & 2.00           & 1.26           & 0.80           & 9.34           & 35.90           & 75.66         \\
Fed-L2P   & 3 (10.00\%)         & 65.00          & 66.95          & 68.83          & 71.33          & 71.64          & 70.17          & 70.29          & 70.58          & 67.02          & 69.78          & 69.16          & \textbf{-4.78}  & 15.84         \\
Fed-DualP & 3 (10.00\%)         & 73.00          & 70.15          & 68.30          & 65.20          & 63.78          & 62.22          & 63.23          & 62.44          & 64.47          & 68.35          & 66.11          & 4.65            & 18.89         \\
PIP-L2P   & 3 (10.00\%)         & 82.60          & 83.35          & 81.20          & 82.95          & 80.24          & 80.17          & 79.04          & 75.99          & 76.36          & 70.39          & 79.23          & 12.21           & 5.77          \\
PIP-DualP & 3 (10.00\%)         & \textbf{90.40} & \textbf{87.10} & \textbf{85.03} & \textbf{86.90} & \textbf{85.92} & \textbf{85.05} & \textbf{83.64} & \textbf{82.64} & \textbf{82.40} & \textbf{80.95} & \textbf{85.00} & 9.45            & \textbf{0.00} \\ \hline
GLFC      & 2 (6.67\%)          & 53.40          & 39.20          & 27.40          & 11.08          & -              & -              & -              & -              & -              & -              & 32.77          & 42.33           & 49.36         \\
LGA       & 2 (6.67\%)          & 71.40          & 59.75          & 57.17          & 54.43          & 49.96          & -              & -              & -              & -              & -              & 58.54          & 21.44           & 23.59         \\
TARGET    & 2 (6.67\%)          & 34.40          & 16.85          & 10.67          & 5.80           & 2.72           & 1.53           & 1.16           & 0.92           & 0.84           & 0.68           & 7.56           & 33.72           & 74.57         \\
Fed-L2P   & 2 (6.67\%)          & 59.00          & 57.60          & 61.87          & 64.88          & 66.52          & 67.37          & 67.74          & 68.30          & 67.62          & 71.57          & 65.25          & \textbf{-12.57} & 16.88         \\
Fed-DualP & 2 (6.67\%)          & 67.50          & 63.95          & 65.50          & 66.43          & 68.30          & 68.90          & 67.66          & 67.25          & 65.37          & 69.43          & 67.03          & -1.93           & 15.10         \\
PIP-L2P   & 2 (6.67\%)          & 83.00          & 80.15          & 78.57          & 80.40          & 77.96          & 77.63          & 75.03          & 71.40          & 69.29          & 68.25          & 76.17          & 14.75           & 5.96          \\
PIP-DualP & 2 (6.67\%)          & \textbf{90.10} & \textbf{84.40} & \textbf{82.97} & \textbf{84.23} & \textbf{82.46} & \textbf{82.33} & \textbf{80.33} & \textbf{78.51} & \textbf{78.61} & \textbf{77.34} & \textbf{82.13} & 12.76           & \textbf{0.00} \\ \hline
\end{tabular}
\caption{Complete numerical results on TinyImageNet (T=10) with smaller participating clients in one-seeded run i.e. 2021. GLFC and LGA performance is averaged from their first 4 and 5 tasks respectively due to crash}
\label{tab:tinyimagenet_nclients}
\end{table*}

\vspace{-5pt}
\section{Complete Numerical Result of Experiment on Smaller Global 
Rounds}\label{app:smaller_rounds}
\vspace{-3pt}
In this section, we present the detailed numerical results of smaller global rounds on CIFAR100, and TinyImageNet datasets. Table \ref{tab:cifar_nrounds}, and \ref{tab:tinyimagenet_nrounds} presents detailed numerical results in CIFAR100, and TinyImageNet respectively with various numbers of global rounds.
\begin{table*}[]
\footnotesize
\centering
\begin{tabular}{lcccccccccccc}
\hline
                   &                                                                & \multicolumn{11}{c}{CIFAR100(T=10)}                                                                                                                                                      \\ \hline
Method             & \begin{tabular}[c]{@{}c@{}}\#Global \\ Rounds (R)\end{tabular} & 10             & 20             & 30             & 40             & 50             & 60             & 70             & 80             & 90             & 100            & \textbf{Avg}   \\ \hline
GLFC(L=10)         & 100                                                            & 90.00          & 82.30          & 77.00          & 72.30          & 65.00          & 66.30          & 59.70          & 56.30          & 50.30          & 50.00          & 66.92          \\
LGA(L=10)          & 100                                                            & 89.60          & 83.20          & 79.30          & 76.10          & 72.90          & 71.70          & 68.40          & 65.70          & 64.70          & 62.90          & 73.45          \\
TARGET(L=10)       & 100                                                            & 65.10          & 37.15          & 26.50          & 21.25          & 19.00          & 16.43          & 14.40          & 13.19          & 11.38          & 10.03          & 23.44          \\
GLFC               & 100                                                            & 86.70          & 74.50          & 71.70          & 65.63          & 65.26          & 60.37          & 54.51          & 52.63          & 48.30          & 44.56          & 62.42          \\
LGA                & 100                                                            & 85.50          & 74.30          & 75.70          & 72.80          & 71.02          & 66.50          & 64.57          & 61.04          & 61.22          & 56.05          & 68.87          \\
TARGET             & 100                                                            & 46.00          & 26.45          & 16.87          & 13.02          & 10.18          & 7.90           & 7.04           & 5.88           & 4.63           & 3.97           & 14.19          \\
Fed-L2P            & 100                                                            & 80.40          & 74.20          & 73.77          & 71.00          & 71.30          & 69.65          & 68.29          & 69.04          & 69.44          & 72.33          & 71.94          \\
Fed-DualP          & 100                                                            & 88.30          & 83.20          & 80.27          & 77.45          & 74.26          & 71.73          & 73.91          & 73.34          & 72.42          & 74.72          & 76.96          \\
\textbf{PIP-L2P}   & 100                                                            & 94.50          & 84.80          & 80.30          & 78.30          & 73.70          & 70.47          & 67.34          & 66.60          & 67.18          & 67.62          & 75.08          \\
\textbf{PIP-DualP} & 100                                                            & \textbf{98.40} & \textbf{92.00} & \textbf{90.57} & \textbf{89.03} & \textbf{87.72} & \textbf{83.30} & \textbf{83.69} & \textbf{82.71} & \textbf{82.36} & \textbf{80.88} & \textbf{87.06} \\ \hline
Fed-L2P            & 80                                                             & 78.80          & 78.30          & 75.40          & 72.10          & 69.98          & 68.92          & 68.67          & 69.79          & 71.89          & 71.83          & 72.57          \\
Fed-DualP          & 80                                                             & 90.10          & 76.95          & 71.93          & 71.40          & 68.14          & 69.55          & 69.14          & 67.66          & 67.93          & 68.92          & 72.17          \\
\textbf{PIP-L2P}   & 80                                                             & 86.70          & 83.05          & 75.57          & 76.05          & 72.98          & 71.85          & 70.24          & 70.15          & 67.49          & 68.09          & 74.22          \\
\textbf{PIP-DualP} & 80                                                             & \textbf{96.70} & \textbf{90.15} & \textbf{88.80} & \textbf{86.70} & \textbf{83.34} & \textbf{79.20} & \textbf{79.34} & \textbf{78.30} & \textbf{77.94} & \textbf{77.28} & \textbf{83.78} \\ \hline
Fed-L2P            & 60                                                             & 79.40          & 70.95          & 72.23          & 73.03          & 71.50          & 70.57          & 70.11          & 69.11          & 68.86          & 67.86          & 71.36          \\
Fed-DualP          & 60                                                             & 85.90          & 85.40          & 81.37          & 78.53          & 74.14          & 72.68          & 73.16          & 72.63          & 73.77          & 73.78          & 77.13          \\
\textbf{PIP-L2P}   & 60                                                             & 92.00          & 78.05          & 72.80          & 70.33          & 67.50          & 65.60          & 66.26          & 65.68          & 65.21          & 64.14          & 70.76          \\
\textbf{PIP-DualP} & 60                                                             & \textbf{97.60} & \textbf{91.40} & \textbf{87.03} & \textbf{86.10} & \textbf{85.36} & \textbf{82.13} & \textbf{82.00} & \textbf{81.69} & \textbf{80.80} & \textbf{80.15} & \textbf{85.43} \\ \hline
Fed-L2P            & 40                                                             & 78.60          & 68.10          & 67.50          & 66.30          & 64.30          & 64.40          & 65.11          & 63.89          & 64.67          & 61.50          & 66.44          \\
Fed-DualP          & 40                                                             & 85.30          & 84.40          & 81.10          & 78.13          & 73.82          & 71.95          & 73.00          & 71.28          & 72.42          & 71.47          & 76.29          \\
\textbf{PIP-L2P}   & 40                                                             & 85.30          & 77.20          & 65.43          & 64.10          & 64.34          & 64.13          & 63.07          & 64.53          & 63.60          & 62.97          & 67.47          \\
\textbf{PIP-DualP} & 40                                                             & \textbf{97.70} & \textbf{93.65} & \textbf{88.13} & \textbf{86.38} & \textbf{84.20} & \textbf{82.97} & \textbf{82.26} & \textbf{81.36} & \textbf{81.27} & \textbf{79.87} & \textbf{85.78} \\ \hline
Fed-L2P            & 20                                                             & 74.20          & 63.55          & 56.07          & 49.75          & 48.62          & 48.15          & 50.17          & 49.51          & 49.61          & 47.52          & 53.72          \\
Fed-DualP          & 20                                                             & 88.20          & 81.45          & 78.07          & 76.45          & 74.10          & 72.60          & 71.06          & 69.98          & 69.34          & 68.91          & 75.02          \\
\textbf{PIP-L2P}   & 20                                                             & 87.60          & 76.95          & 71.57          & 64.40          & 64.12          & 62.82          & 63.01          & 61.98          & 60.29          & 57.75          & 67.05          \\ 
\textbf{PIP-DualP} & 20                                                             & \textbf{97.80} & \textbf{91.05} & \textbf{86.40} & \textbf{83.33} & \textbf{83.42} & \textbf{81.27} & \textbf{81.13} & \textbf{82.01} & \textbf{81.01} & \textbf{80.28} & \textbf{84.77} \\ \hline
\end{tabular}
\caption{Complete numerical results on CIFAR100 (T=10, local clients = 3) with smaller rounds in one-seeded run i.e. 2021}
\label{tab:cifar_nrounds}
\end{table*}

\begin{table*}[]
\footnotesize
\centering
\begin{tabular}{lcccccccccccc}
\hline
                   &                                                                & \multicolumn{11}{c}{TinyImageNet(T=10)}                                                                                                                                                  \\ \hline
Method             & \begin{tabular}[c]{@{}c@{}}\#Global \\ Rounds (R)\end{tabular} & 20             & 40             & 60             & 80             & 100            & 120            & 140            & 160            & 180            & 200            & \textbf{Avg}   \\ \hline
GLFC(L=10)         & 100                                                            & 66.00          & 58.30          & 55.30          & 51.00          & 47.70          & 45.30          & 43.00          & 40.00          & 37.30          & 35.00          & 47.89          \\
LGA(L=10)          & 100                                                            & 70.30          & 64.00          & 60.30          & 58.00          & 55.80          & 53.10          & 47.90          & 45.30          & 39.80          & 37.30          & 53.18          \\
TARGET(L=10)       & 100                                                            & 56.30          & 33.20          & 21.00          & 17.18          & 11.88          & 7.77           & 7.01           & 4.94           & 2.99           & 2.02           & 16.43          \\
GLFC               & 100                                                            & 55.60          & 37.70          & 26.37          & 19.88          & -              & -              & -              & -              & -              & -              & 34.89          \\
LGA                & 100                                                            & 73.30          & 69.55          & 60.30          & 52.75          & 47.16          & -              & -              & -              & -              & -              & 60.61          \\
TARGET             & 100                                                            & 36.70          & 20.20          & 12.37          & 8.60           & 5.06           & 3.53           & 2.89           & 2.00           & 1.26           & 0.80           & 9.34           \\
Fed-L2P            & 100                                                            & 65.00          & 66.95          & 68.83          & 71.33          & 71.64          & 70.17          & 70.29          & 70.58          & 67.02          & 69.78          & 69.16          \\
Fed-DualP          & 100                                                            & 86.60          & 75.70          & 71.33          & 66.33          & 64.28          & 62.83          & 62.27          & 61.19          & 59.99          & 62.60          & 67.31          \\
\textbf{PIP-L2P}   & 100                                                            & 82.60          & 83.35          & 81.20          & 82.95          & 80.24          & 80.17          & 79.04          & 75.99          & 76.36          & 70.39          & 79.23          \\ \hline
\textbf{PIP-DualP} & 100                                                            & \textbf{92.70} & \textbf{87.90} & \textbf{86.83} & \textbf{88.23} & \textbf{87.04} & \textbf{87.18} & \textbf{85.96} & \textbf{85.13} & \textbf{84.62} & \textbf{82.35} & \textbf{86.79} \\
Fed-L2P            & 80                                                             & 65.00          & 68.55          & 71.20          & 71.30          & 71.12          & 71.87          & 73.04          & 73.66          & 72.28          & 73.03          & 71.10          \\
Fed-DualP          & 80                                                             & 74.80          & 72.05          & 71.60          & 70.08          & 68.14          & 66.72          & 66.63          & 65.80          & 65.57          & 71.81          & 69.32          \\
\textbf{PIP-L2P}   & 80                                                             & 82.20          & 83.15          & 82.37          & 83.65          & 82.00          & 80.18          & 78.39          & 77.03          & 76.82          & 74.39          & 80.02          \\
\textbf{PIP-DualP} & 80                                                             & \textbf{90.10} & \textbf{87.10} & \textbf{87.17} & \textbf{87.53} & \textbf{86.16} & \textbf{86.33} & \textbf{85.23} & \textbf{83.98} & \textbf{83.06} & \textbf{81.64} & \textbf{85.83} \\ \hline
Fed-L2P            & 60                                                             & 66.20          & 70.30          & 71.77          & 74.25          & 72.90          & 73.77          & 74.47          & 73.05          & 72.36          & 71.15          & 72.02          \\
Fed-DualP          & 60                                                             & 73.40          & 73.10          & 72.07          & 70.60          & 69.30          & 67.68          & 66.69          & 67.60          & 66.43          & 71.02          & 69.79          \\
\textbf{PIP-L2P}   & 60                                                             & 79.70          & 79.70          & 80.60          & 80.20          & 79.50          & 79.43          & 75.39          & 73.98          & 72.87          & 70.13          & 77.15          \\
\textbf{PIP-DualP} & 60                                                             & \textbf{91.50} & \textbf{87.20} & \textbf{87.97} & \textbf{89.30} & \textbf{87.76} & \textbf{87.78} & \textbf{86.20} & \textbf{85.88} & \textbf{84.78} & \textbf{82.94} & \textbf{87.13} \\ \hline
Fed-L2P            & 40                                                             & 70.20          & 72.95          & 72.53          & 74.28          & 74.58          & 73.78          & 73.33          & 71.13          & 71.03          & 70.37          & 72.42          \\
Fed-DualP          & 40                                                             & 72.30          & 72.05          & 73.27          & 71.60          & 71.48          & 69.70          & 68.74          & 67.20          & 67.26          & 69.40          & 70.30          \\
\textbf{PIP-L2P}   & 40                                                             & 80.20          & 78.60          & 77.67          & 79.78          & 78.10          & 76.27          & 75.87          & 72.15          & 71.92          & 70.42          & 76.10          \\
\textbf{PIP-DualP} & 40                                                             & \textbf{90.70} & \textbf{87.15} & \textbf{87.17} & \textbf{87.43} & \textbf{86.98} & \textbf{86.75} & \textbf{85.17} & \textbf{84.05} & \textbf{83.26} & \textbf{82.11} & \textbf{86.08} \\ \hline
Fed-L2P            & 20                                                             & 61.00          & 59.45          & 63.23          & 64.85          & 65.10          & 65.03          & 62.39          & 60.35          & 59.19          & 57.15          & 61.77          \\
Fed-DualP          & 20                                                             & 73.90          & 75.55          & 75.10          & 74.28          & 72.44          & 72.75          & 71.51          & 69.45          & 68.78          & 70.71          & 72.45          \\
\textbf{PIP-L2P}   & 20                                                             & 80.20          & 78.15          & 77.43          & 77.83          & 76.12          & 76.85          & 74.14          & 73.64          & 71.77          & 67.18          & 75.33          \\
\textbf{PIP-DualP} & 20                                                             & \textbf{89.20} & \textbf{87.90} & \textbf{87.63} & \textbf{88.48} & \textbf{87.52} & \textbf{86.68} & \textbf{84.86} & \textbf{83.00} & \textbf{83.31} & \textbf{82.14} & \textbf{86.07} \\ \hline
\end{tabular}
\caption{Complete numerical results on TinyImageNet (T=10, local clients = 3) with smaller rounds in one-seeded run i.e. 2021}
\label{tab:tinyimagenet_nrounds}
\end{table*}

%%%%%%%%%%%%%%%%%%%%%%%%%%%%%%%%%%%%%%%%%%%%%%%%%%%%%%%%%

% \bibliography{samples/sample-base}
% \bibliographystyle{ACM-Reference-Format}
% \bibliography{main}

\end{document}